\documentclass{article}

\PassOptionsToPackage{numbers, compress}{natbib}
 \usepackage[preprint]{neurips_2026}

\usepackage[utf8]{inputenc} % allow utf-8 input
\usepackage[T1]{fontenc}    % use 8-bit T1 fonts
\usepackage{hyperref}       % hyperlinks
\usepackage{url}            % simple URL typesetting
\usepackage{booktabs}       % professional-quality tables
\usepackage{amsfonts}       % blackboard math symbols
\usepackage{nicefrac}       % compact symbols for 1/2, etc.
\usepackage{microtype}      % microtypography
\usepackage{xcolor}         % colors
\usepackage{multirow}

\usepackage{graphicx} % 用于 \resizebox 调整表格自适应页宽
\usepackage{subcaption}
\usepackage{amsmath}
\usepackage{cleveref}
\usepackage{wrapfig}

\usepackage[export]{adjustbox}

\title{HiTokSR: A Coarse-to-Fine Tokenizer with Hierarchical Codebooks for High-Fidelity Real-World Image Super-Resolution}

\author{%
  Mingxi Li\\
  \\
  \texttt{li\_mx\_0318@163.com}
}

\begin{document}

\maketitle

\begin{abstract}
Vector-quantized (VQ) generative models have shown promising results in real-world image super-resolution (Real-ISR). However, existing methods typically rely on a monolithic latent space that entangles low-frequency structures with high-frequency textures. This entanglement forces a single codebook to capture a combinatorially complex set of structure-texture pairings, which constrains representational capacity and limits codebook utilization. To address this issue, we present HiTokSR, a hierarchical token prediction framework. Instead of using a single codebook, HiTokSR partitions the latent space along the channel dimension into frequency-aware groups, quantizing each with an independent sub-codebook. This coarse-to-fine design disentangles global structures from fine details, enhancing combinatorial expressiveness while circumventing the optimization instability of high-dimensional nearest-neighbor lookups. To further improve semantic consistency, our generator integrates priors from a vision foundation model via adaptive feature modulation, multi-scale class tokens, and a representation alignment loss. Additionally, we introduce an index-level perturbation strategy during decoder fine-tuning to bridge the train-test discrepancy in discrete token prediction. Extensive experiments on real-world benchmarks demonstrate that HiTokSR achieves state-of-the-art performance in both perceptual quality and reconstruction fidelity.

% This paper presents HiTokSR, a hierarchical token prediction framework for real-world image super-resolution that revisits codebook design in vector-quantized (VQ) generation. Existing VQ-based methods typically operate on a monolithic latent space, where low-frequency structures and high-frequency textures are entangled within a single codebook. This entanglement forces the codebook to capture a combinatorially large set of structure–texture pairings, which constrains representational capacity without necessarily improving codebook utilization. HiTokSR instead partitions the latent space along the channel dimension into frequency-aware groups, each quantized with an independent sub-codebook. This coarse-to-fine design separates global structure from fine detail, enabling combinatorial expressiveness while circumventing the optimization instability caused by high-dimensional nearest-neighbor lookup. To further improve semantic consistency, the generator is conditioned on priors from a vision foundation model through adaptive feature modulation, multi-scale class tokens, and a representation alignment loss. A decoder fine-tuning stage with index-level perturbation is introduced to close the train–test discrepancy in discrete token prediction. Experiments on real-world benchmarks show that HiTokSR achieves state-of-the-art perceptual quality and reconstruction fidelity.

\end{abstract}
\section{Introduction}

Real-world image super-resolution (SR) aims to reconstruct high-quality images from low-resolution inputs degraded by complex and unknown processes. Early generative approaches based on Generative Adversarial Networks (GANs)~\cite{ESRGAN, realesrgan, BSRGAN, LDL} demonstrated the ability to synthesize sharp details through adversarial training, but they often suffer from training instability and produce unnatural artifacts or overly smooth textures in the face of complex real-world degradations. More recently, diffusion models~\cite{StableSR, DiffBIR, SeeSR, Resshift} have achieved strong perceptual quality through iterative denoising. However, their multi-step inference incurs substantial computational cost~\cite{OSEDiff, adcsr, TSDSR, TinySR}, and the stochastic sampling process can introduce semantic drift~\cite{FaithDiff, pisasr}, where generated content deviates from the structural context of the low-resolution input. These drawbacks have spurred interest in efficient single-pass alternatives. Vector-quantized (VQ)~\cite{vqvae, vqgan} methods offer such an alternative by predicting discrete tokens in one forward pass, but their reconstruction quality is limited by the design of the underlying codebook.
\begin{figure*}[!htb]
    \centering
    %================ (a) 第一行：Coarse to Fine ================
    \begin{minipage}{\textwidth}
        \centering
        \begin{subfigure}[b]{0.2\textwidth}  % ← 改为 0.19
            \centering
            \setlength{\lineskip}{0pt}
            \subcaption*{$M=1$}
            \includegraphics[width=\linewidth]{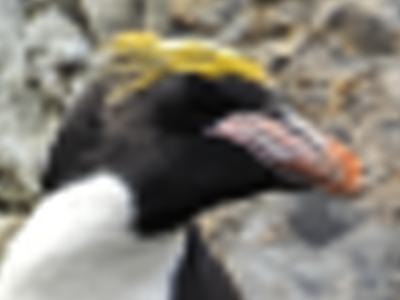}
            \includegraphics[width=\linewidth]{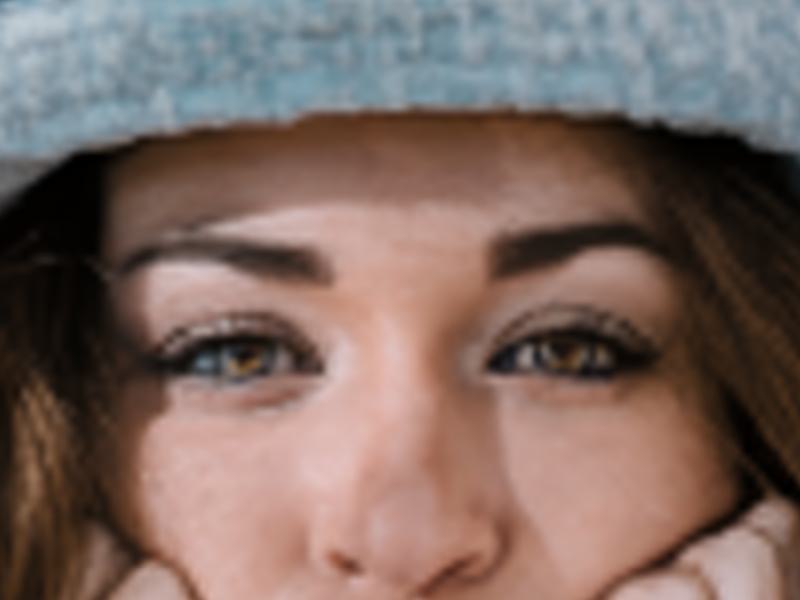}
        \end{subfigure}%
        \begin{subfigure}[b]{0.2\textwidth}
            \centering
            \setlength{\lineskip}{0pt}
            \subcaption*{$M=2$}
            \includegraphics[width=\linewidth]{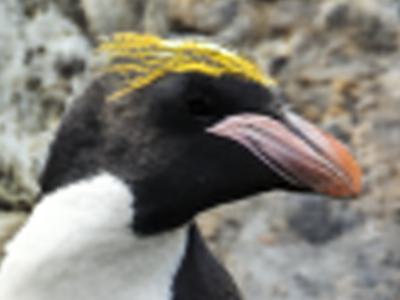}
            \includegraphics[width=\linewidth]{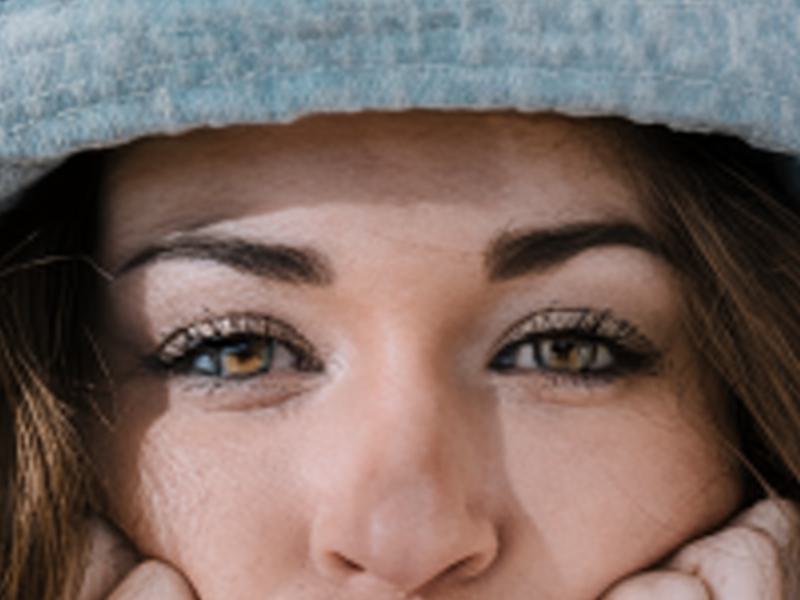}
        \end{subfigure}%
        \begin{subfigure}[b]{0.2\textwidth}
            \centering
            \setlength{\lineskip}{0pt}
            \subcaption*{$M=3$}
            \includegraphics[width=\linewidth]{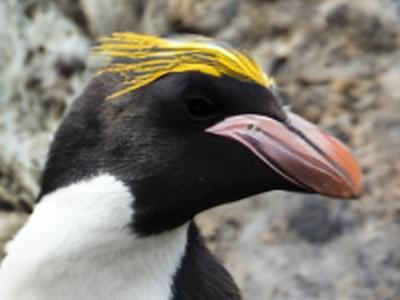}
            \includegraphics[width=\linewidth]{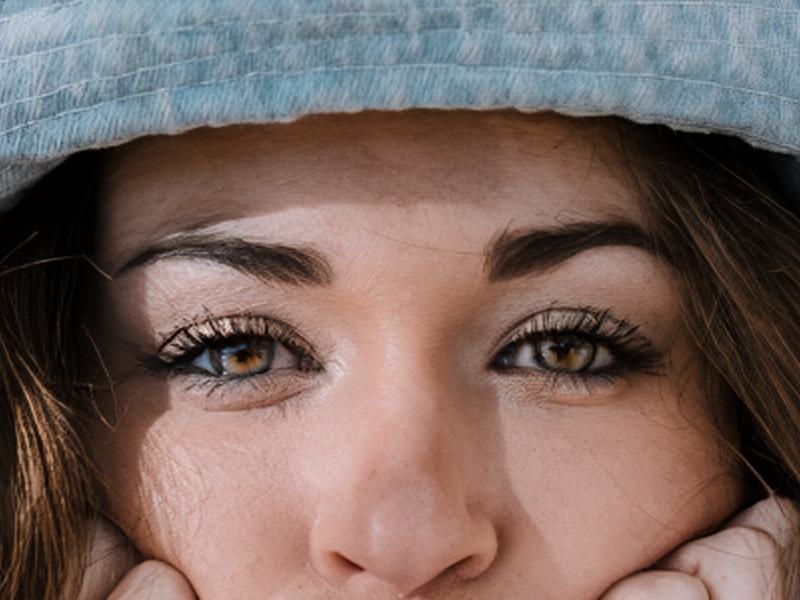}
        \end{subfigure}%
        \begin{subfigure}[b]{0.2\textwidth}
            \centering
            \setlength{\lineskip}{0pt}
            \subcaption*{$M=4$}
            \includegraphics[width=\linewidth]{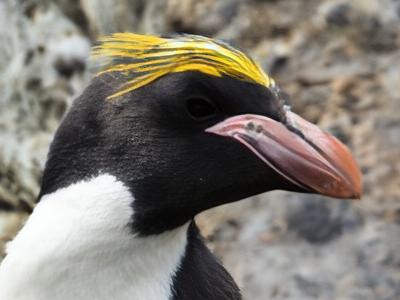}
            \includegraphics[width=\linewidth]{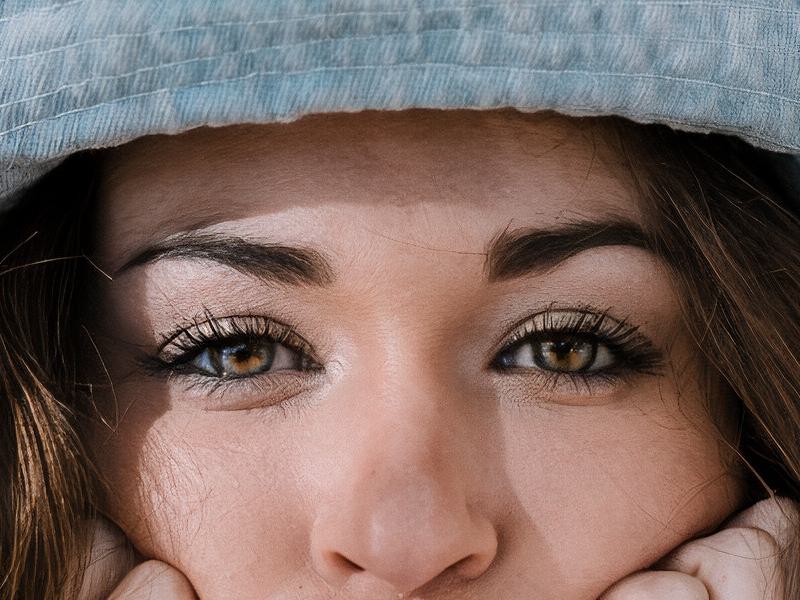}
        \end{subfigure}%
        \begin{subfigure}[b]{0.2\textwidth}
            \centering
            \setlength{\lineskip}{0pt}
            \subcaption*{GT}
            \includegraphics[width=\linewidth]{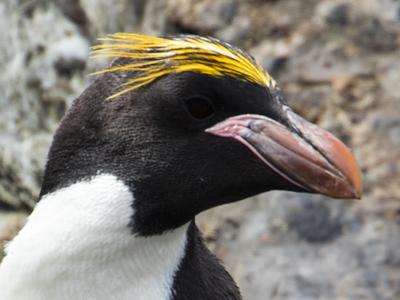}
            \includegraphics[width=\linewidth]{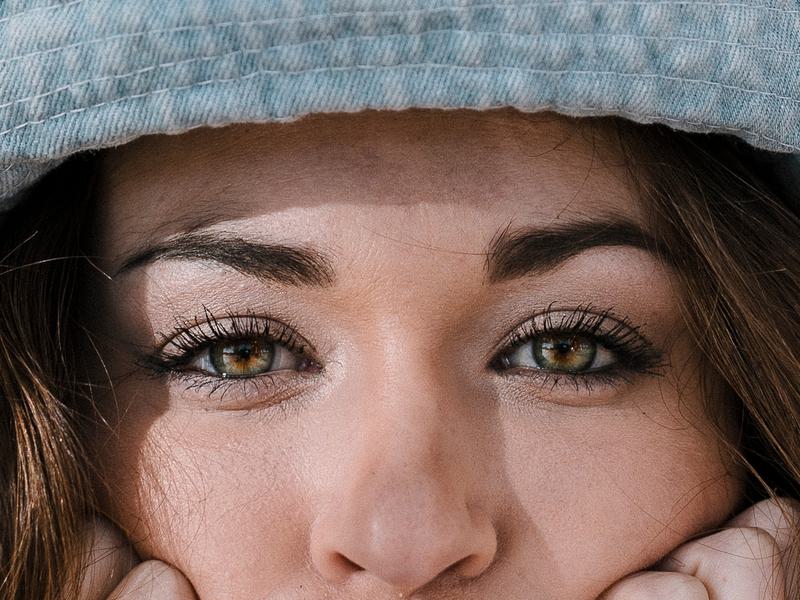}
        \end{subfigure}
        \subcaption{Coarse-to-fine hierarchical reconstruction. \textbf{M} denotes number of active groups.}
        \label{fig:intro_init_a}
    \end{minipage}

    % 可选分割线（当前已注释）
    % \par\smallskip
    % \textcolor{gray!50}{\rule{\textwidth}{0.4pt}}
    % \par\smallskip

    %================ (b) 第二行左侧：对比图 ================
    %================ (c) 第二行右侧：效率散点图 ================
    \begin{minipage}[t]{0.625\textwidth}
        \centering
        \begin{subfigure}[b]{0.25\linewidth}
            \centering
            \setlength{\lineskip}{0pt}
            \includegraphics[width=\linewidth]{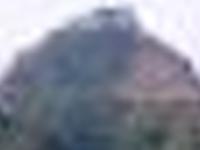}
            \includegraphics[width=\linewidth]{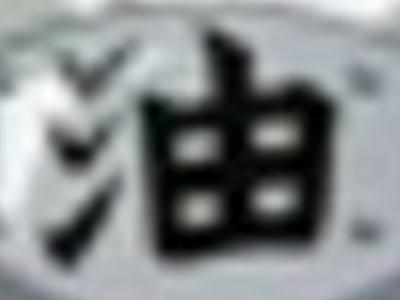}
            \subcaption*{LQ}
        \end{subfigure}%
        \begin{subfigure}[b]{0.25\linewidth}
            \centering
            \setlength{\lineskip}{0pt}
            \includegraphics[width=\linewidth]{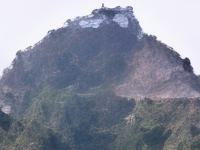}
            \includegraphics[width=\linewidth]{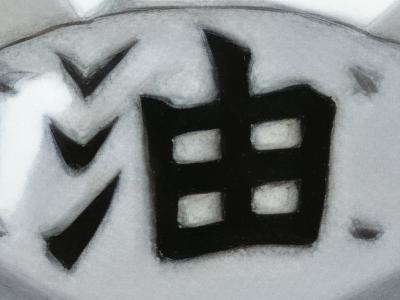}
            \subcaption*{OSEDiff}
        \end{subfigure}%
        \begin{subfigure}[b]{0.25\linewidth}
            \centering
            \setlength{\lineskip}{0pt}
            \includegraphics[width=\linewidth]{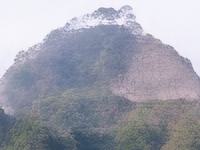}
            \includegraphics[width=\linewidth]{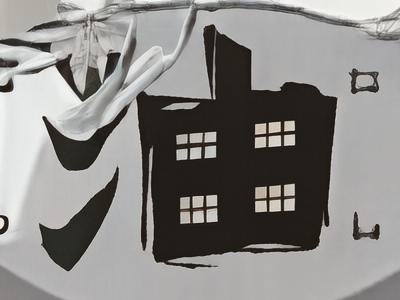}
            \subcaption*{VARSR}
        \end{subfigure}%
        \begin{subfigure}[b]{0.25\linewidth}
            \centering
            \setlength{\lineskip}{0pt}
            \includegraphics[width=\linewidth]{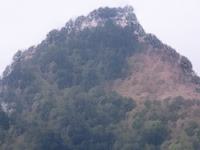}
            \includegraphics[width=\linewidth]{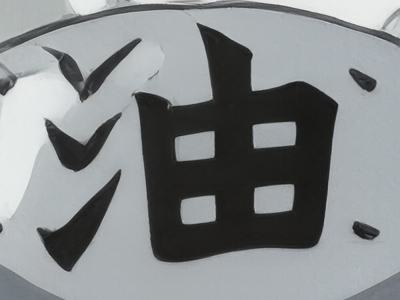}
            \subcaption*{Ours}
        \end{subfigure}
        \subcaption{Visual comparison on real-world examples.}
        \label{fig:intro_init_b}
    \end{minipage}%
    \begin{minipage}[t]{0.375\textwidth}
        \centering
        \includegraphics[width=\linewidth]{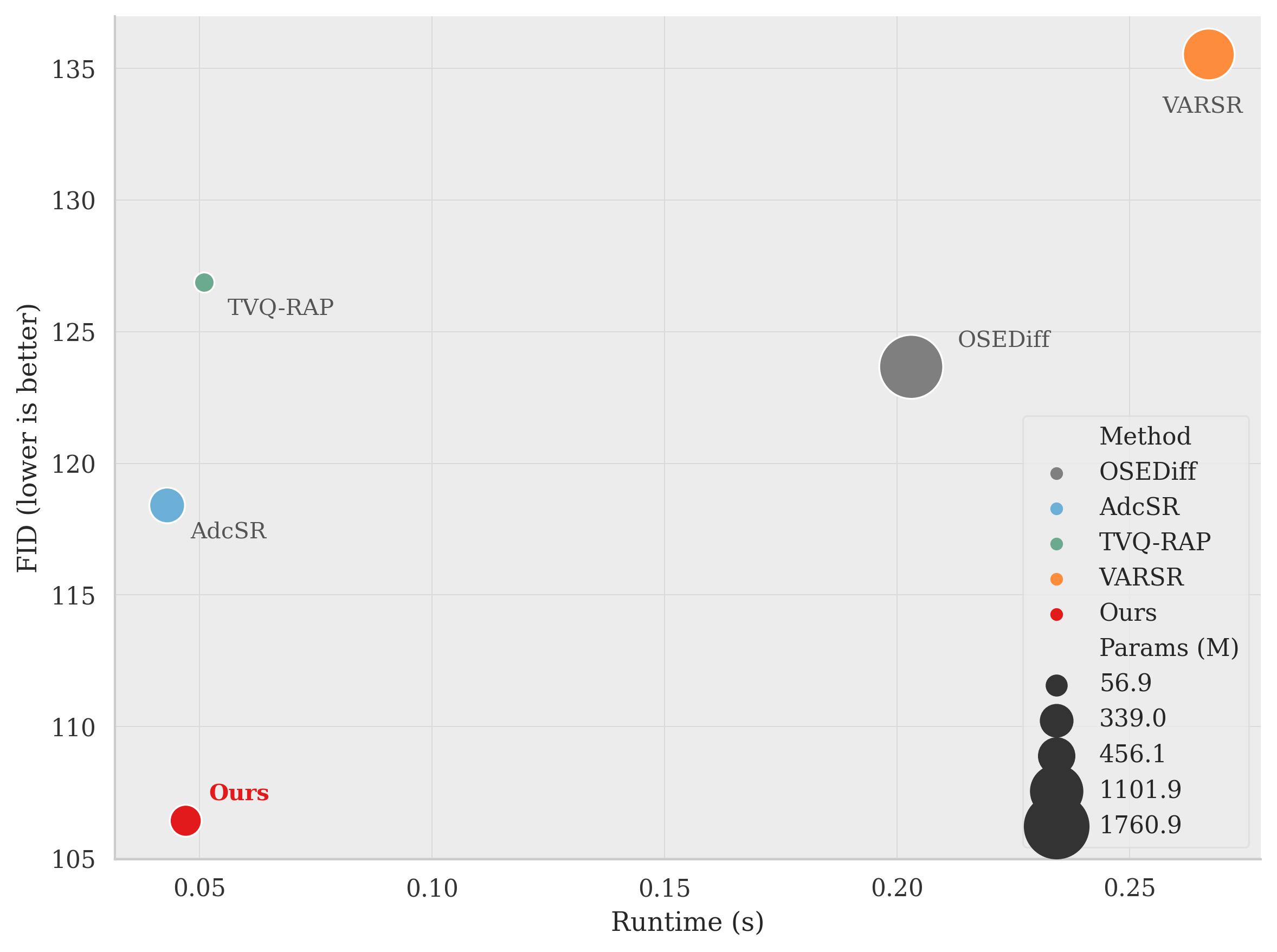}
        \subcaption{Efficiency vs. performance on RealSR.}
        \label{fig:intro_init_c}
    \end{minipage}
    \caption{HiTokSR leverages coarse-to-fine hierarchical tokenization to achieve a competitive balance between inference efficiency and perceptual quality.
    (a) Unlike monolithic quantizers that entangle structures and textures, our tokenizer splits the latent space into channel groups, which learns a coarse-to-fine hierarchy. Early groups capture global structure, and later groups add fine details as \textbf{M} increases.
    (b) This design produces more realistic high-resolution outputs, with textures that compare favorably against recent diffusion-based and VQ-based methods.
    (c) As a single-pass prediction framework, HiTokSR achieves competitive perceptual quality at substantially lower computational cost, resulting in a practical efficiency-quality trade-off.}
    \label{fig:intro_init}
\end{figure*}

% Most existing VQ-based SR methods, including CodeFormer~\cite{Codeformer} and FeMaSR~\cite{FeMaSR}, adopt a monolithic latent space where a single codebook jointly encodes all spatial and spectral information. We identify two limitations of this design that compound each other. The first is an optimization difficulty: expanding the codebook size or latent dimension to increase capacity renders high-dimensional nearest-neighbor lookup increasingly unreliable due to the curse of dimensionality, which in turn leads to low codebook utilization and diminishing returns~\cite{vqgan-lc, rqvae, VAR}. This makes it difficult to scale monolithic codebooks in practice. The second is a structural inefficiency: even at a given codebook size, entangling low-frequency structures and high-frequency textures forces the codebook to capture a combinatorially large set of structure–texture pairings. This is inherently wasteful, since a single structural pattern typically corresponds to numerous textural variations. Together, these two limitations create a self-reinforcing bottleneck: scaling the codebook to increase expressiveness is blocked by the curse of dimensionality, while operating at a manageable size leaves the entanglement problem unaddressed.

Most existing VQ-based SR methods, including CodeFormer~\cite{Codeformer} and FeMaSR~\cite{FeMaSR}, adopt a monolithic latent space where a single codebook jointly encodes all spatial and spectral information. While straightforward, this design suffers from two fundamental limitations that compound each other. The first is an optimization difficulty: expanding the codebook size or latent dimension to increase representational capacity renders high-dimensional nearest-neighbor lookup increasingly unreliable due to the curse of dimensionality. As the dimension grows, the distance between any two points becomes noisy, leading to low codebook utilization and diminishing returns~\cite{vqgan-lc, rqvae, VAR}. In practice, many codebook entries are left unused or updated infrequently, wasting capacity and destabilizing training. This makes it difficult to scale monolithic codebooks effectively. The second is a structural inefficiency: even at a manageable codebook size, entangling low-frequency structures and high-frequency textures within a single codeword forces the codebook to capture a combinatorially large set of structure-texture pairings. This is inherently wasteful, since a single structural pattern, such as an edge or a smooth region, typically corresponds to numerous possible textural variations in natural images. Memorizing all such pairings exhaustively is impractical. Together, these two limitations create a self-reinforcing bottleneck: scaling the codebook to increase expressiveness is blocked by the curse of dimensionality, while operating at a manageable size leaves the entanglement problem unaddressed.

We propose HiTokSR, a framework built around a hierarchical codebook design that addresses both limitations simultaneously. Instead of quantizing the latent space as a whole, HiTokSR partitions it along the channel dimension into multiple non-overlapping groups, each quantized with an independent sub-codebook. Because each sub-codebook operates in a lower-dimensional subspace, nearest-neighbor lookup remains stable and reliable, effectively circumventing the optimization difficulties that plague large monolithic codebooks. At the same time, this decomposition naturally allows different sub-codebooks to specialize in distinct frequency bands: early groups can focus on capturing global structures, while later groups concentrate on fine details. As a result, the overall discrete representation can express a wide variety of structure-texture combinations without requiring any single codebook to memorize an exhaustive set of pairings (Figure~\ref{fig:intro_init}(a)). To explicitly encourage this spectral specialization, we introduce a group masking strategy combined with frequency-matched supervision via discrete wavelet transform, which guides early groups toward coarse structures and later groups toward fine textures in a principled manner. %(~\Cref{fig:intro_init_a})

Building on this tokenizer, the generation process incorporates two additional components. First, while the hierarchical codebook provides an efficient discrete representation, the generator must still infer high-level scene content from a degraded input, where complex degradations often obscure object semantics and structural layout. We therefore condition the Transformer~\cite{Transformer} generator on semantic priors from a vision foundation model~\cite{CLIP, Siglip2, dinov3} through adaptive feature modulation and multi-scale class tokens, with a representation alignment loss that encourages the semantic features of degraded inputs to match those of clean targets. Second, a train–test discrepancy arises because the decoder is trained on ground-truth tokens but receives predicted tokens at inference. We mitigate this by fine-tuning the decoder under controlled index-level perturbation, sampling from the top-K predictions rather than using the top-1 token, which improves robustness to prediction errors without modifying the generator.

Experiments on real-world benchmarks demonstrate that HiTokSR achieves state-of-the-art perceptual quality and reconstruction fidelity. Notably, our method attains these results while maintaining the efficient single-pass inference characteristic of VQ-based frameworks, offering a practical balance between generation quality and computational cost (Figures~\ref{fig:intro_init}(b) and Figures~\ref{fig:intro_init}(c)). %(~\Cref{fig:intro_init_b,fig:intro_init_c}).

\section{Related Work}
\label{sec:related_work}

\subsection{Real-World Image Super-Resolution}
Real-world image super-resolution aims to restore high-quality images from low-resolution inputs that are degraded by complex and unknown processes, including noise, blur, and compression artifacts. Early data-driven methods employed Generative Adversarial Networks (GANs)~\cite{SRGAN, ESRGAN, realesrgan, BSRGAN, LDL, SPSR, DualFormer} to enhance perceptual quality. However, they often suffered from training instability and limited reconstruction fidelity. Diffusion models~\cite{StableSR, DiffBIR, SeeSR, Resshift, Dreamclear} have advanced the state of the art through iterative denoising, achieving strong perceptual quality but incurring substantial computational overhead and a risk of stochastic semantic drift~\cite{FaithDiff, pisasr}. While one-step diffusion-based super-resolution methods~\cite{OSEDiff, adcsr, TSDSR, TinySR} reduce the number of sampling steps, they introduce a trade-off between efficiency and perceptual quality. Vector-quantized (VQ) approaches have emerged as an effective alternative by predicting discrete tokens in a single forward pass~\cite{Codeformer, FeMaSR, TVQ}. Despite offering considerable computational advantages over diffusion models, existing VQ-based methods remain constrained by the representational capacity of their codebook designs. The conflict between expanding codebook capacity and maintaining optimization stability continues to pose a fundamental challenge, highlighting the need for more effective codebook designs in this context.

\subsection{Vector Quantization and Codebook Design}
Vector quantization (VQ)~\cite{vqvae, vqgan} maps continuous features to discrete tokens via nearest-neighbor lookup in a learned codebook. Standard VQ suffers from a fundamental trade-off between capacity and stability: increasing the codebook size or embedding dimension leads to low utilization and diminishing returns, as high-dimensional nearest-neighbor search becomes unreliable. Residual quantization (RQ)~\cite{rqvae} and VQGAN-LC~\cite{vqgan-lc} mitigate this through multi-stage decomposition and improved initialization, respectively, while Visual AutoRegressive (VAR)~\cite{VAR, VARSR} reformulates autoregressive generation as coarse-to-fine next-scale prediction.
Despite these advances, most methods still encode different frequency components into a single monolithic vector, forcing the codebook to memorize a large set of structure-texture pairings. The work closest to ours, TVQ-RAP~\cite{TVQ}, separates the latent into continuous structure and discrete texture branches. However, its structure extraction relies on a manually chosen downsampling scale which introduces aliasing and sub-pixel misalignment, causing topological inconsistencies and artifacts. Moreover, its texture branch still uses a monolithic codebook with high channel dimensionality, leaving the core optimization challenges of nearest-neighbor lookup unresolved. In contrast, our tokenizer partitions the latent space along the channel dimension into frequency-aware groups, each quantized with an independent, lower-dimensional sub-codebook. This design aligns with the frequency-selective characteristics of human perception and achieves combinatorial expressiveness while avoiding the optimization challenges associated with large unified codebooks.

% Despite these advances, existing methods encode low-frequency and high-frequency components into a single vector. This design forces the codebook to memorize all possible combinations of structures and textures, which is inherently inefficient because a single low-frequency pattern typically corresponds to numerous high-frequency variations. In contrast, the proposed tokenizer explicitly partitions the latent space along the channel dimension into frequency-aware groups, each with an independent sub-codebook, thereby enabling the free combination of low-frequency and high-frequency content during decoding. This design aligns with the frequency-selective characteristics of human perception and achieves combinatorial capacity while avoiding the optimization challenges associated with large unified codebooks.

\begin{figure}[t]
    \centering
    \includegraphics[width=\textwidth]{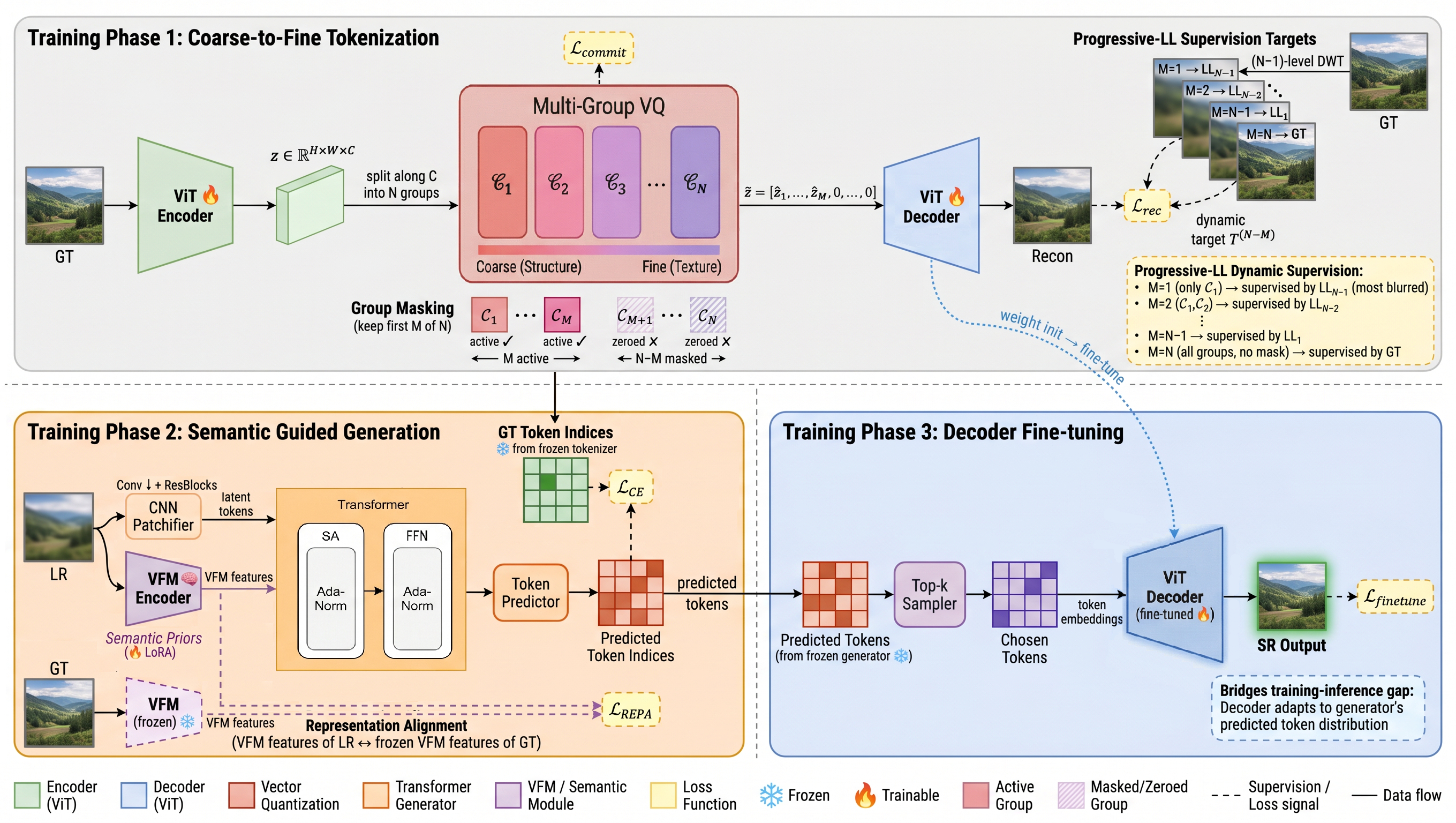}
    \caption{Overview of the proposed HiTokSR framework. For simplicity, the reconstruction loss in Phase 2, which requires passing tokens through the frozen decoder, is omitted from the diagram.}
    \label{fig:overview_of_framework}
\end{figure}

\section{Method}
\label{sec:method}

HiTokSR follows a tokenize-then-generate paradigm. As illustrated in \Cref{fig:overview_of_framework}, we first train a coarse-to-fine tokenizer (\Cref{sec:tokenizer}) that decomposes an image into hierarchical discrete tokens through multi-group vector quantization. We then train a bidirectional Transformer generator (\Cref{sec:generator}) that predicts token indices from a degraded low-resolution input, conditioned on semantic priors from a vision foundation model. Finally, we fine-tune the decoder (\Cref{sec:decoder}) to bridge the distribution gap between ground-truth and predicted tokens.

\subsection{Coarse-to-Fine Image Tokenization}
\label{sec:tokenizer}

Standard vector quantization maps a continuous latent representation to a single discrete codebook via nearest-neighbor lookup. This monolithic design imposes two related burdens. First, scaling the codebook size or latent dimension makes high-dimensional nearest-neighbor lookup unreliable. Second, encoding all frequency components into a single codeword entangles low-frequency structures with high-frequency textures. We propose a hierarchical tokenization scheme that partitions the latent space into multiple frequency-aware groups, each quantized independently in a lower-dimensional subspace.

\paragraph{Grouped Vector Quantization}
\label{sec:grouped_vq}

Given a high-resolution image $\mathbf{x} \in \mathbb{R}^{H \times W \times 3}$, a ViT-based encoder $\mathcal{E}$ extracts continuous latent features $\mathbf{z} = \mathcal{E}(\mathbf{x}) \in \mathbb{R}^{h \times w \times C}$. We partition $\mathbf{z}$ uniformly along the channel dimension into $N$ non-overlapping groups:
\begin{equation}
\mathbf{z} = [\mathbf{z}_1, \mathbf{z}_2, \ldots, \mathbf{z}_N], \quad \mathbf{z}_i \in \mathbb{R}^{h \times w \times (C/N)}. \label{eq:partition}
\end{equation}
The channel dimension $C$ is chosen to be divisible by $N$. Each group $\mathbf{z}_i$ is quantized with a dedicated sub-codebook $\mathcal{C}_i \in \mathbb{R}^{K \times (C/N)}$, where $K$ is the codebook size shared across groups:
\begin{equation}
\hat{\mathbf{z}}_i = \mathcal{C}_i[\mathbf{k}_i], \quad \mathbf{k}_i = \underset{k \in \{1, \ldots, K\}}{\arg\min} \|\mathbf{z}_i - \mathcal{C}_i[k]\|_2. \label{eq:quantize}
\end{equation}
Here $\mathbf{k}_i \in \{1, \ldots, K\}^{h \times w}$ is the discrete token indices for group $i$. The quantized groups are concatenated to form $\hat{\mathbf{z}} = [\hat{\mathbf{z}}_1, \ldots, \hat{\mathbf{z}}_N] \in \mathbb{R}^{h \times w \times C}$. Because each sub-codebook operates in a $C/N$-dimensional subspace rather than the full $C$-dimensional space, nearest-neighbor lookup remains reliable, circumventing the curse of dimensionality that limits monolithic codebooks.

\paragraph{Group Masking Strategy}
\label{sec:masking}

To encourage each group to specialize in a specific frequency band without imposing explicit spectral constraints, we randomly mask a suffix of the groups during training. The number of active groups $M$ is sampled from a heavy-tailed distribution:
\begin{equation}
M = \begin{cases}
N, & \text{with probability } 0.9, \\
m, & \text{with probability } \frac{0.1}{N-1}, \; \text{for } m \in \{1, 2, \ldots, N-1\}.
\end{cases} \label{eq:mask_dist}
\end{equation}
Given a sampled $M$, the masked representation $\tilde{\mathbf{z}}$ is constructed as:
\begin{equation}
\tilde{\mathbf{z}}_i = \begin{cases}
\hat{\mathbf{z}}_i, & \text{if } i \leq M, \\
\mathbf{0}, & \text{if } i > M.
\end{cases} \label{eq:mask}
\end{equation}
This heavy-tailed distribution ensures that the model predominantly learns full-fidelity reconstruction (90\% of steps use all groups), which maintains reconstruction quality at full capacity. In the remaining 10\% of steps, only a prefix of the groups is active. In those steps, the early groups, which are always active, are forced to capture the information necessary for coarse reconstruction, while later groups learn to encode the residual detail needed when they become active.

\paragraph{Frequency-Matched Supervision}
\label{sec:freq_supervision}

We couple the masking strategy with a supervision signal that explicitly matches the frequency content determined by the number of active groups. We perform an $(N-1)$-level discrete wavelet transform (DWT) on the ground-truth image $\mathbf{x}$, producing a sequence of low-frequency approximations $\mathbf{T}^{(0)}, \mathbf{T}^{(1)}, \ldots, \mathbf{T}^{(N-1)}$. Here $\mathbf{T}^{(0)} = \mathbf{x}$ is the original image, and $\mathbf{T}^{(l)}$ is the approximation after $l$ decomposition levels, containing progressively coarser structures as $l$ increases. The reconstruction target is dynamically selected as:
\begin{equation}
\mathbf{T}_{\text{target}} = \mathbf{T}^{(N-M)}. \label{eq:target}
\end{equation}
This establishes a direct one-to-one mapping between the number of active groups and the frequency level of the target. When $M=1$, the target is $\mathbf{T}^{(N-1)}$, the coarsest approximation, so the model learns to reconstruct only the lowest-frequency structures from a single group. As $M$ increases, higher-frequency components are progressively introduced into both the active representation and the supervision signal. This paired design---the mask determines which groups contribute, and the DWT level determines what frequency content is required---guides the tokenizer toward a coarse-to-fine spectral organization without any explicit frequency assignment to individual groups.

\paragraph{Training Objectives}
\label{sec:tokenizer_loss}

The tokenizer is trained with a reconstruction loss and a commitment loss. The reconstruction loss $\mathcal{L}_{\text{rec}}$ is computed on the decoded masked representation against the target $\mathbf{T}_{\text{target}}$ using a combination of L1, LPIPS, and GAN losses:
\begin{equation}
\mathcal{L}_{\text{rec}} = \lambda_{\text{L1}} \|\mathcal{D}(\tilde{\mathbf{z}}) - \mathbf{T}_{\text{target}}\|_1 + \lambda_{\text{LPIPS}} \mathcal{L}_{\text{LPIPS}}(\mathcal{D}(\tilde{\mathbf{z}}), \mathbf{T}_{\text{target}}) + \lambda_{\text{GAN}} \mathcal{L}_{\text{GAN}}(\mathcal{D}(\tilde{\mathbf{z}})), \label{eq:rec_loss}
\end{equation}
where $\lambda_{\text{L1}}, \lambda_{\text{LPIPS}}, \lambda_{\text{GAN}}$ are scalar weights. Quantization is performed before masking, so the commitment loss applies to all $N$ groups regardless of which are active in the current step:
\begin{equation}
\mathcal{L}_{\text{commit}} = \sum_{i=1}^{N} \left( \|\mathrm{sg}[\mathbf{z}_i] - \hat{\mathbf{z}}_i\|_2^2 + \beta \|\mathbf{z}_i - \mathrm{sg}[\hat{\mathbf{z}}_i]\|_2^2 \right), \label{eq:commit_loss}
\end{equation}
where $\mathrm{sg}[\cdot]$ denotes the stop-gradient operator and $\beta$ is a hyperparameter. The first term moves each codebook vector toward the encoder output, and the second term encourages the encoder to commit to its nearest codebook entry. Applying the commitment loss to all groups at every step ensures that all sub-codebooks receive gradient updates, preventing the collapse of groups that happen to be masked in the current iteration. The total tokenizer loss is:
\begin{equation}
\mathcal{L}_{\text{total}} = \mathcal{L}_{\text{rec}} + \lambda \mathcal{L}_{\text{commit}}, \label{eq:total_tokenizer}
\end{equation}
where $\lambda$ balances reconstruction fidelity and codebook commitment.

\subsection{Semantic-Guided Token Generation}
\label{sec:generator}

Once the tokenizer is trained, the super‑resolution task reduces to predicting the token indices \(\{\mathbf{k}_i\}_{i=1}^{N}\) from a degraded low‑resolution input. We train a generator with a hybrid CNN‑Transformer architecture for this task: a convolutional patchifier first extracts spatial features and projects them into a sequence of latent tokens, which are then processed by a stack of Transformer blocks with RMSNorm, SwiGLU activation, and 2D rotary positional embeddings.

To predict semantically plausible tokens, the generator must resolve not only local texture details but also high‑level scene semantics that determine which textures are appropriate in a given context. The degraded input alone often lacks sufficient information to disambiguate such cases, for instance, distinguishing similar textures on semantically different surfaces. We therefore incorporate semantic priors from a vision foundation model (VFM). The VFM encoder produces features that are robust to common degradations, encoding stable information about object semantics, geometric layout, and material properties.

We integrate VFM priors at two levels. First, we extract the final‑layer patch tokens from the VFM encoder. These spatially dense features are used to predict channel‑wise scale and shift parameters that modulate the features within each Transformer block through adaptive normalization. After RMSNorm, the features \(\mathbf{f}\) are modulated as
\begin{equation}
\mathbf{f}' = \gamma(\mathbf{f}_{\text{VFM}}) \odot \mathrm{RMSNorm}(\mathbf{f}) + \beta(\mathbf{f}_{\text{VFM}}), \label{eq:ada_norm}
\end{equation}
where \(\gamma(\cdot)\) and \(\beta(\cdot)\) are small projection heads. This allows the generator to dynamically adjust its representations based on fine‑grained semantic content. Second, we concatenate multi‑scale CLS tokens extracted from different layers of the VFM encoder into the generator’s input sequence. During self‑attention, queries from local patches can attend to these global tokens, which serve as compressed semantic summaries of the scene across multiple receptive fields. The CLS tokens are discarded after the final Transformer block, adding negligible overhead.

The VFM encoder is pretrained on high‑quality images and may produce suboptimal features for degraded inputs. We therefore fine‑tune it using LoRA to adapt to the training degradation distribution while preserving pretrained knowledge. To bridge the semantic gap between degraded and clean features, we apply a Representation Alignment (REPA) loss between VFM features of the low‑quality input and those of the ground‑truth image, encouraging degradation‑robust representations aligned with clean semantics.

\paragraph{Training Objective}
The generator is optimized with token‑level cross‑entropy loss and pixel‑level reconstruction losses. The cross‑entropy loss compares predicted token probabilities against ground‑truth token indices from the pretrained tokenizer. Pixel‑level L1, LPIPS, and GAN losses are applied to the decoded output, and gradients are propagated back to the generator via the Straight‑Through Estimator to bypass the non‑differentiable token selection. The overall objective is
\begin{equation}
\mathcal{L}_{\text{gen}} = \lambda_{\text{L1}}\mathcal{L}_{\text{L1}} + \lambda_{\text{CE}} \mathcal{L}_{\text{CE}} + \lambda_{\text{REPA}} \mathcal{L}_{\text{REPA}} + \lambda_{\text{LPIPS}} \mathcal{L}_{\text{LPIPS}} + \lambda_{\text{GAN}} \mathcal{L}_{\text{GAN}}, \label{eq:gen_loss}
\end{equation}
where the \(\lambda\) coefficients balance each term.

\subsection{Decoder Fine-Tuning}
\label{sec:decoder}

The generator described above is trained with the decoder frozen. This creates a train–test discrepancy: the decoder is trained exclusively on ground-truth tokens from the tokenizer, yet at inference it receives predicted tokens whose distribution inevitably differs due to generator errors. This discrepancy is compounded by the Straight-Through Estimator (STE) used for generator training. Since discrete token selection is non-differentiable, pixel-level losses must back-propagate through a quantization step approximated by STE. The resulting approximate gradients provide only coarse feedback about how token predictions affect reconstruction quality, which can contribute to a distribution gap between the tokens seen by the decoder during training and those produced at inference.

To bridge this gap, we propose a decoder fine-tuning stage after generator training is finished. Both the generator and the tokenizer's sub-codebooks remain frozen and only the decoder is updated. The core idea is to directly expose the decoder to the token prediction errors it will face at inference. Instead of using the single top-$1$ token, we stochastically sample the input token from the top-$K$ candidates of the generator's softmax output at each spatial position. This perturbed token stream acts as an online augmentation, forcing the decoder to learn a mapping from a neighborhood of plausible tokens to the ground-truth image. Consequently, the decoder becomes robust to inaccuracies in the generator's predictions.

We supervise this stage with a combination of L1, LPIPS, and GAN losses. At inference, we revert to deterministic top-$1$ selection. The stochastic fine-tuning ensures that the decoder reliably tolerates the prediction errors encountered in practice.

\begin{table*}[t]
\centering
\caption{Quantitative comparison on real-world benchmarks. The best and second-best values for each metric are highlighted in \textcolor{red}{red} and \textcolor{blue}{blue}, respectively.}
\label{tab:quantitative_comp}
\resizebox{0.95\linewidth}{!}{%
{\footnotesize
\setlength{\tabcolsep}{2.2pt}
\renewcommand{\arraystretch}{1.05}
\begin{tabular}{ll|ccc|cccccc|ccc|c}
\toprule
\textbf{Datasets} & \textbf{Metrics}
& \textbf{StableSR} & \textbf{DiffBIR} & \textbf{SeeSR}
& \textbf{SinSR} & \textbf{OSEDiff} & \textbf{AdcSR} & \textbf{TSD-SR} & \textbf{FaithDiff} & \textbf{PiSA-SR}
& \textbf{FeMASR} & \textbf{TVQ-RAP} & \textbf{VARSR}
& \textbf{HiTokSR} \\
& \textbf{Steps}
& 200 & 50 & 50
& 1 & 1 & 1 & 1 & 1 & 1
& 1 & 1 & 10
& 1 \\
\midrule

\multirow{9}{*}{RealSR}
& PSNR$\uparrow$
& 25.51 & 24.83 & 25.30
& \textcolor{blue}{26.15} & 25.15 & 25.47 & 24.81 & 25.24 & 25.50
& 25.06 & 24.44 & 25.47
& \textcolor{red}{26.16} \\
& SSIM$\uparrow$
& \textcolor{red}{0.7491} & 0.6501 & 0.7265
& 0.7370 & 0.7341 & 0.7301 & 0.7172 & 0.7049 & \textcolor{blue}{0.7418}
& 0.7356 & 0.7070 & 0.7213
& 0.7403 \\
& LPIPS$\downarrow$
& \textcolor{blue}{0.2604} & 0.3650 & 0.2988
& 0.3068 & 0.2921 & 0.2885 & 0.2743 & 0.2906 & 0.2672
& 0.2937 & 0.2937 & 0.3308
& \textcolor{red}{0.2477} \\
& DISTS$\downarrow$
& \textcolor{blue}{0.1990} & 0.2399 & 0.2227
& 0.2332 & 0.2127 & 0.2129 & 0.2105 & 0.2138 & 0.2044
& 0.2286 & 0.2176 & 0.2406
& \textcolor{red}{0.1975} \\
& FID$\downarrow$
& 132.07 & 130.76 & 124.01
& 137.20 & 123.66 & 118.40 & 114.46 & \textcolor{blue}{111.16} & 124.16
& 141.07 & 126.86 & 135.52
& \textcolor{red}{106.43} \\
& CLIPIQA$\uparrow$
& 0.5424 & 0.7053 & 0.6721
& 0.6320 & 0.6679 & 0.6729 & \textcolor{red}{0.7159} & 0.6132 & 0.6700
& 0.5407 & 0.6827 & \textcolor{blue}{0.7057}
& 0.6924 \\
& NIQE$\downarrow$
& 6.63 & 5.84 & 5.30
& 6.07 & 5.64 & 5.35 & \textcolor{blue}{5.12} & 5.37 & 5.51
& 5.77 & \textcolor{red}{4.96} & 6.06
& 5.13 \\
& MUSIQ$\uparrow$
& 61.81 & 69.28 & 70.04
& 61.36 & 69.08 & 69.90 & \textcolor{blue}{71.18} & 68.56 & 70.14
& 59.06 & 66.38 & \textcolor{red}{71.46}
& 65.90 \\
& MANIQA$\uparrow$
& 0.5950 & 0.6502 & 0.6465
& 0.5414 & 0.6335 & 0.6353 & 0.6346 & \textcolor{red}{0.6719} & 0.6551
& 0.4862 & 0.5903 & \textcolor{blue}{0.6571}
& 0.6069 \\

\midrule

\multirow{9}{*}{DRealSR}
& PSNR$\uparrow$
& \textcolor{blue}{29.02} & 25.90 & 28.11
& 28.18 & 27.92 & 28.10 & 27.77 & 28.37 & 28.32
& 26.87 & 26.05 & 28.01
& \textcolor{red}{29.19} \\
& SSIM$\uparrow$
& \textcolor{red}{0.8044} & 0.6245 & 0.7660
& 0.7529 & 0.7835 & 0.7726 & 0.7559 & 0.7335 & 0.7804
& 0.7570 & 0.7101 & 0.7585
& \textcolor{blue}{0.7895} \\
& LPIPS$\downarrow$
& \textcolor{blue}{0.2698} & 0.4669 & 0.3193
& 0.3517 & 0.2967 & 0.3046 & 0.2966 & 0.3286 & 0.2960
& 0.3157 & 0.3656 & 0.3633
& \textcolor{red}{0.2691} \\
& DISTS$\downarrow$
& \textcolor{blue}{0.2066} & 0.2882 & 0.2327
& 0.2468 & 0.2163 & 0.2200 & 0.2136 & 0.2308 & 0.2169
& 0.2239 & 0.2474 & 0.2570
& \textcolor{red}{0.2050} \\
& FID$\downarrow$
& 151.26 & 180.37 & 153.59
& 172.07 & 135.59 & 134.03 & 135.31 & 147.12 & \textcolor{blue}{130.47}
& 157.83 & 172.62 & 157.13
& \textcolor{red}{127.16} \\
& CLIPIQA$\uparrow$
& 0.4907 & 0.7071 & 0.6889
& 0.6635 & 0.6960 & 0.7050 & \textcolor{red}{0.7348} & 0.6134 & 0.6969
& 0.5636 & 0.7108 & \textcolor{blue}{0.7239}
& 0.7012 \\
& NIQE$\downarrow$
& 7.54 & 6.33 & 6.43
& 6.70 & 6.44 & 6.45 & 5.91 & 6.20 & 6.18
& 5.91 & \textcolor{red}{5.46} & 6.85
& \textcolor{blue}{5.82} \\
& MUSIQ$\uparrow$
& 51.36 & 66.14 & 65.14
& 57.27 & 64.70 & 66.26 & \textcolor{blue}{66.60} & 62.86 & 66.10
& 53.71 & 63.98 & \textcolor{red}{68.65}
& 60.90 \\
& MANIQA$\uparrow$
& 0.4965 & \textcolor{red}{0.6221} & 0.6094
& 0.5032 & 0.5898 & 0.5915 & 0.5873 & 0.6158 & 0.6161
& 0.4392 & 0.5587 & \textcolor{blue}{0.6220}
& 0.5704 \\

\midrule

\multirow{7}{*}{RealLQ250}
& CLIPIQA$\uparrow$
& 0.5095 & 0.7136 & 0.7036
& 0.7165 & 0.6723 & 0.6888 & \textcolor{blue}{0.7369} & 0.6629 & 0.7055
& 0.6216 & 0.7329 & \textcolor{red}{0.7535}
& 0.7285 \\
& NIQE$\downarrow$
& 4.69 & 5.12 & 4.41
& 5.43 & 3.97 & 3.72 & \textcolor{blue}{3.70} & 4.13 & 3.92
& 4.30 & 4.26 & 5.11
& \textcolor{red}{3.45} \\
& MUSIQ$\uparrow$
& 56.84 & 67.53 & 70.50
& 65.41 & 69.56 & 69.98 & \textcolor{red}{73.22} & 70.03 & 71.24
& 61.85 & 70.18 & \textcolor{blue}{72.85}
& 68.42 \\
& MANIQA$\uparrow$
& 0.5098 & 0.5877 & 0.5935
& 0.5253 & 0.5783 & 0.5815 & 0.5924 & \textcolor{red}{0.6368} & \textcolor{blue}{0.6053}
& 0.4924 & 0.5849 & 0.6052
& 0.5828 \\
& Q-Insight$\uparrow$
& 1.91 & 1.92 & 1.71
& \textcolor{blue}{1.99} & 1.98 & 1.74 & \textcolor{blue}{1.99} & 1.50 & 1.96
& 1.90 & \textcolor{red}{2.18} & 1.88
& 1.94 \\
% & UniPercept\_IAA$\uparrow$
% & 55.58 & 61.79 & 63.63
% & 62.28 & 63.59 & 64.35 & \textcolor{red}{65.75} & \textcolor{blue}{65.12} & 62.47
% & 56.30 & 62.27 & 64.34
% & 61.62 \\
& UniPercept\_IQA$\uparrow$
& 57.36 & 65.22 & 68.90
& 62.37 & 69.00 & 67.65 & \textcolor{red}{70.30} & \textcolor{blue}{70.01} & 68.31
& 58.82 & 65.57 & 67.80
& 66.73 \\
% & UniPercept\_ISTA$\uparrow$
% & 39.51 & 40.05 & 40.65
% & 41.23 & 39.97 & 39.67 & 41.34 & 40.62 & 40.10
% & 40.18 & \textcolor{blue}{41.62} & \textcolor{red}{42.21}
% & 40.52 \\
& VisualQuality-R1$\uparrow$
& 3.85 & 4.28 & 4.46
& 3.84 & \textcolor{red}{4.56} & 4.49 & 4.47 & 4.49 & \textcolor{blue}{4.51}
& 3.84 & 4.15 & 4.42
& 4.47 \\

\bottomrule
\end{tabular}
}
}
\end{table*}

\section{Experiments}
\label{sec:experiments}

\subsection{Experimental Settings}
\paragraph{Datasets and Metrics}
Following prior studies, we build the training set by merging the LSDIR~\cite{LSDIR} dataset with the first 10K face images from FFHQ~\cite{FFHQ}. To synthesize realistic low-resolution counterparts, we adopt the degradation pipeline of Real-ESRGAN. All training pairs are generated with a scaling factor of $\times 4$. For evaluation, we employ three real-world benchmarks RealSR~\cite{RealSR}, DRealSR~\cite{DRealSR} and RealLQ250~\cite{Dreamclear}. RealLQ250 lacks corresponding GT images. We evaluate performance from both fidelity and perceptual perspectives. Full-reference metrics include PSNR, SSIM~\cite{SSIM}, DISTS~\cite{DISTS}, LPIPS~\cite{LPIPS}, and FID~\cite{FID}. For no-reference quality evaluation, we adopt MUSIQ~\cite{MUSIQ}, MANIQA~\cite{MANIQA}, CLIPIQA~\cite{CLIPIQA}, NIQE~\cite{NIQE}, as well as the recently proposed Q-Insight~\cite{QInsight}, UniPercept-IQA~\cite{Unipercept} and VisualQuality-R1~\cite{Visualqualityr1}, covering a wide range of blind perceptual assessment.

\paragraph{Comparison Methods}
We compare our method with a representative set of generative super-resolution frameworks, including both diffusion/flow-based models (StableSR~\cite{StableSR}, DiffBIR~\cite{DiffBIR}, SeeSR~\cite{SeeSR}, SinSR~\cite{SinSR}, OSEDiff~\cite{OSEDiff}, AdcSR~\cite{adcsr}, TSD-SR~\cite{TSDSR}, FaithDiff~\cite{FaithDiff} and PiSA-SR~\cite{pisasr}) and VQ-based approaches (FeMASR~\cite{FeMaSR}, TVQ-RAP~\cite{TVQ} and VARSR~\cite{VARSR}).

% \paragraph{Evaluation Metrics.} 

Other implementation details are provided in the \textbf{Appendix}.

\subsection{Comparison with Other Methods}
\paragraph{Quantitative Comparisons.}
\Cref{tab:quantitative_comp} reports a comprehensive comparison on three real-world benchmarks: RealSR, DRealSR, and RealLQ250, covering multi-step diffusion models, one-step diffusion models, and VQ-based approaches.
On RealSR and DRealSR, HiTokSR attains the best LPIPS, DISTS, FID, and PSNR among all compared methods, with particularly clear margins in LPIPS and FID. On RealLQ250, which evaluates no-reference perceptual metrics exclusively, HiTokSR achieves the best NIQE and ranks third in CLIPIQA behind VARSR and TSD-SR.
While no single method dominates every metric, HiTokSR achieves more balanced performance across fidelity-oriented and perceptual quality metrics. Notably, this balanced profile is achieved with a fraction of the computational cost required by most diffusion-based alternatives, as discussed in the efficiency comparisons below.

\begin{figure*}[t]
    \centering
    \begin{subfigure}[c]{0.13\textwidth}
        \centering
        \includegraphics[width=\linewidth]{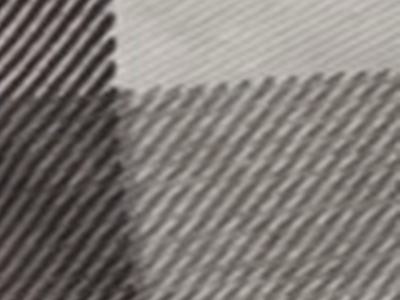}
        \subcaption*{LQ}
    \end{subfigure}
    \begin{subfigure}[c]{0.13\textwidth}
        \centering
        \includegraphics[width=\linewidth]{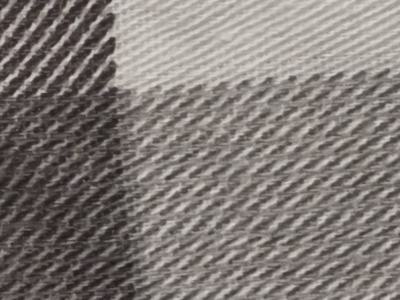}
        \subcaption*{HR}
    \end{subfigure}
    \begin{subfigure}[c]{0.13\textwidth}
        \centering
        \includegraphics[width=\linewidth]{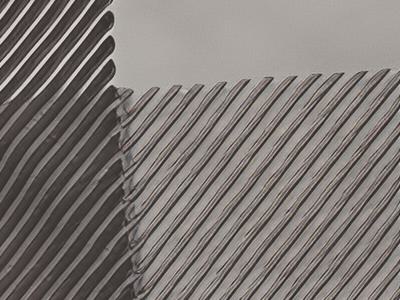}
        \subcaption*{StableSR}
    \end{subfigure}
    \begin{subfigure}[c]{0.13\textwidth}
        \centering
        \includegraphics[width=\linewidth]{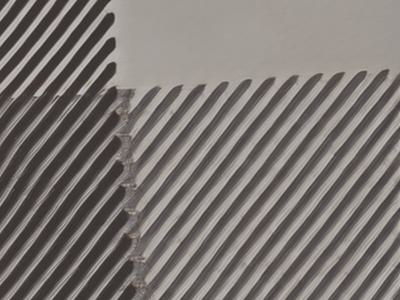}
        \subcaption*{OSEDiff}
    \end{subfigure}
    \begin{subfigure}[c]{0.13\textwidth}
        \centering
        \includegraphics[width=\linewidth]{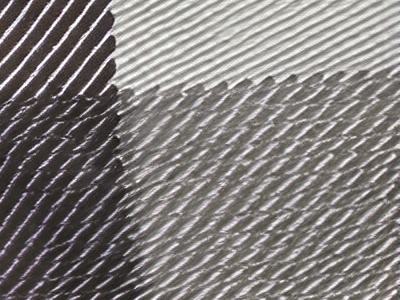}
        \subcaption*{TVQ-RAP}
    \end{subfigure}
    \begin{subfigure}[c]{0.13\textwidth}
        \centering
        \includegraphics[width=\linewidth]{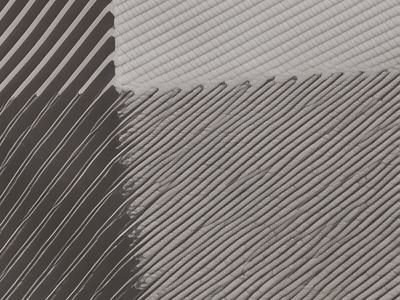}
        \subcaption*{VARSR}
    \end{subfigure}
    \begin{subfigure}[c]{0.13\textwidth}
        \centering
        \includegraphics[width=\linewidth]{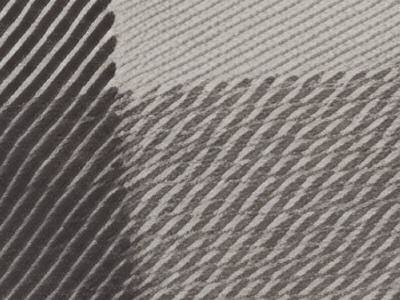}
        \subcaption*{HiTokSR}
    \end{subfigure}

    \begin{subfigure}[c]{0.13\textwidth}
        \centering
        \includegraphics[width=\linewidth]{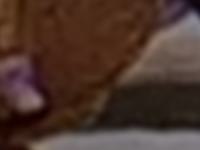}
        \subcaption*{LQ}
    \end{subfigure}
    \begin{subfigure}[c]{0.13\textwidth}
        \centering
        \includegraphics[width=\linewidth]{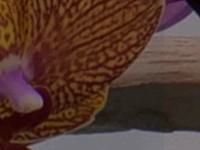}
        \subcaption*{HR}
    \end{subfigure}
    \begin{subfigure}[c]{0.13\textwidth}
        \centering
        \includegraphics[width=\linewidth]{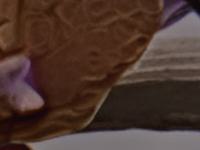}
        \subcaption*{StableSR}
    \end{subfigure}
    \begin{subfigure}[c]{0.13\textwidth}
        \centering
        \includegraphics[width=\linewidth]{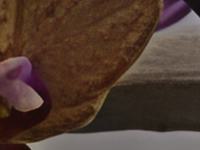}
        \subcaption*{OSEDiff}
    \end{subfigure}
    \begin{subfigure}[c]{0.13\textwidth}
        \centering
        \includegraphics[width=\linewidth]{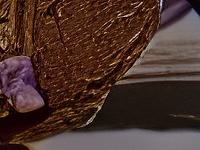}
        \subcaption*{TVQ-RAP}
    \end{subfigure}
    \begin{subfigure}[c]{0.13\textwidth}
        \centering
        \includegraphics[width=\linewidth]{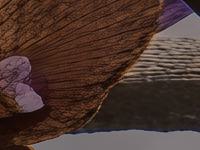}
        \subcaption*{VARSR}
    \end{subfigure}
    \begin{subfigure}[c]{0.13\textwidth}
        \centering
        \includegraphics[width=\linewidth]{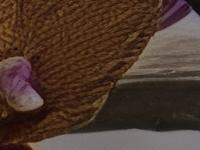}
        \subcaption*{HiTokSR}
    \end{subfigure}

    \begin{subfigure}[c]{0.13\textwidth}
        \centering
        \includegraphics[width=\linewidth]{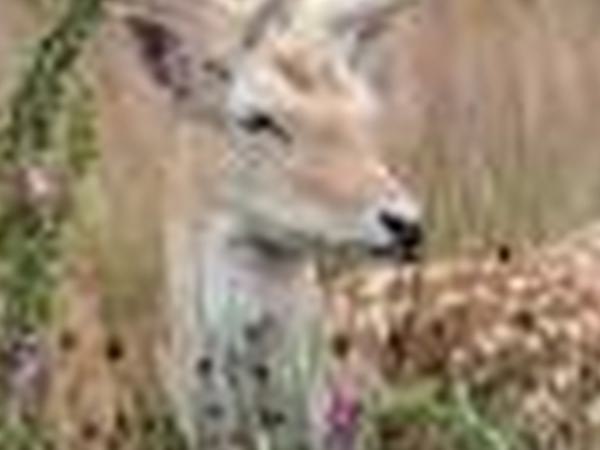}
        \subcaption*{LQ}
    \end{subfigure}
    \begin{subfigure}[c]{0.13\textwidth}
        \centering
        \includegraphics[width=\linewidth]{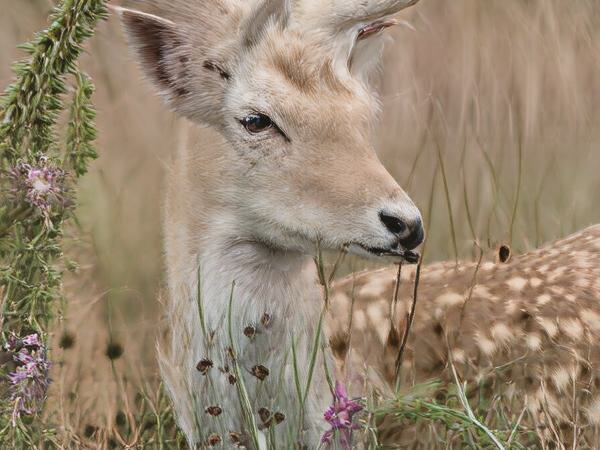}
        \subcaption*{TSD-SR}
    \end{subfigure}
    \begin{subfigure}[c]{0.13\textwidth}
        \centering
        \includegraphics[width=\linewidth]{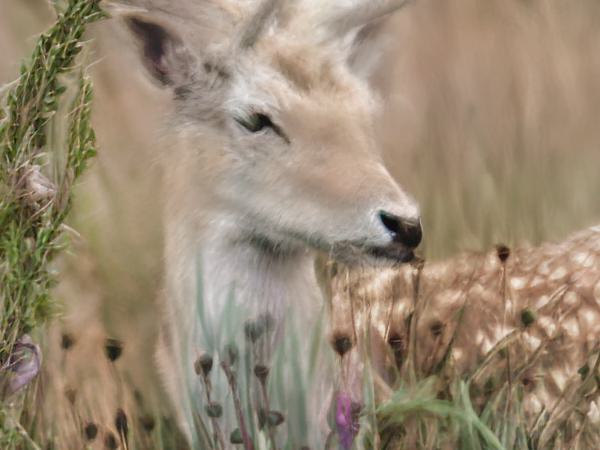}
        \subcaption*{StableSR}
    \end{subfigure}
    \begin{subfigure}[c]{0.13\textwidth}
        \centering
        \includegraphics[width=\linewidth]{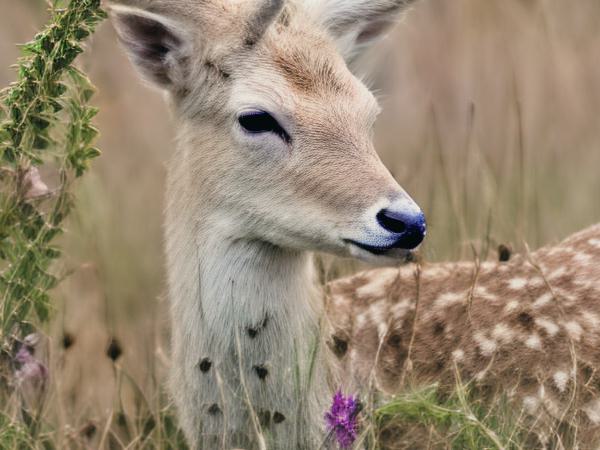}
        \subcaption*{OSEDiff}
    \end{subfigure}
    \begin{subfigure}[c]{0.13\textwidth}
        \centering
        \includegraphics[width=\linewidth]{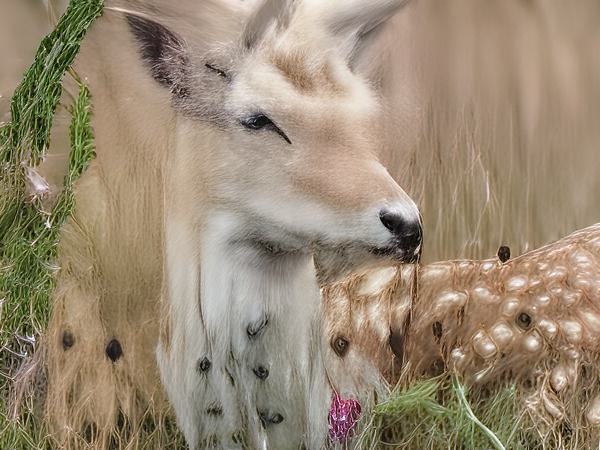}
        \subcaption*{TVQ-RAP}
    \end{subfigure}
    \begin{subfigure}[c]{0.13\textwidth}
        \centering
        \includegraphics[width=\linewidth]{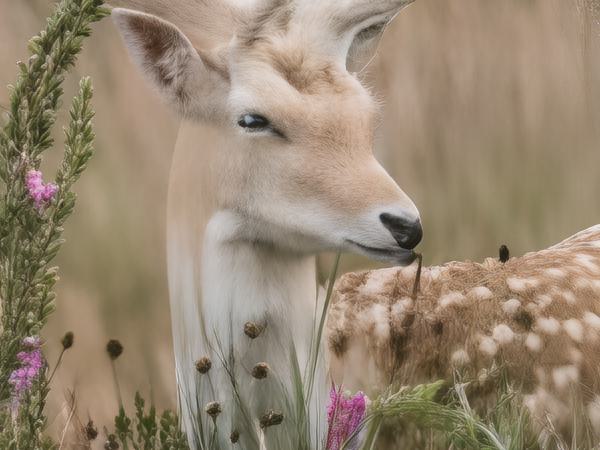}
        \subcaption*{VARSR}
    \end{subfigure}
    \begin{subfigure}[c]{0.13\textwidth}
        \centering
        \includegraphics[width=\linewidth]{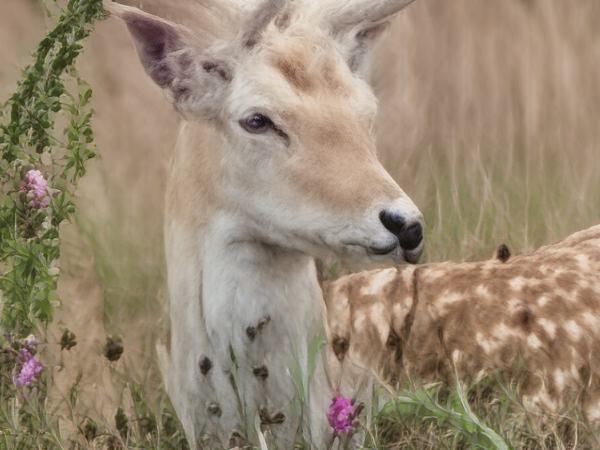}
        \subcaption*{HiTokSR}
    \end{subfigure}
    
    \caption{Qualitative comparisons with Real-SR methods on three real-world benchmarks: RealSR (1st row), DRealSR (2nd row) and RealLQ250 (3rd row, GT unavailable). Best viewed zoomed in.}
    \label{fig:sec4_visual_comp}
\end{figure*}

\paragraph{Qualitative Comparisons.}
We compares restoration quality across representative examples in \Cref{fig:sec4_visual_comp}. While existing methods typically trade off between over-smoothing and artifact introduction, HiTokSR produces more realistic outputs by disentangling structure and texture through its coarse-to-fine tokenizer.
On \textbf{structured high-frequency textures} such as fabric stripes, diffusion-based methods either over-smooth details or introduce sharpening artifacts, while VARSR lacks textural fidelity. HiTokSR restores natural patterns that closely match the ground truth.
On \textbf{fine-grained natural textures} such as leaf surfaces, leaf vein structures are easily lost or hallucinated. TSD-SR and StableSR blur these delicate patterns; VARSR produces sharp yet structurally implausible textures. HiTokSR recovers rich hierarchical detail with greater faithfulness and a more natural appearance.
For \textbf{complex multi-semantic scenes}, competing methods often produce over-smoothed fur or hallucinated background details. In contrast, HiTokSR restores notably natural fur and eye details while limiting implausible background textures.

\begin{table}[t]
    \centering
    \caption{Comparison of inference efficiency and perceptual quality on the RealSR dataset. Efficiency metrics are benchmarked with $128\times128$ input images on a single NVIDIA 4090 GPU.}
    \resizebox{0.9\linewidth}{!}{%
    \begin{tabular}{c | c c c c | c c c}
        \toprule
        Methods & Runtime (s) & MACs (G) & Params. (M) & Memory (GiB) & LPIPS$\downarrow$ & FID$\downarrow$ & CLIPIQA$\uparrow$ \\
        \midrule
        OSEDiff & 0.203 & 2339.8 & 1760.9 & 3.75 & 0.2921 & 123.66 & 0.6679 \\
        AdcSR & \textcolor{red}{0.043} & \textcolor{red}{496.0} & 456.1 & 2.19 & \textcolor{blue}{0.2885} & \textcolor{blue}{118.40} & 0.6729 \\
        TVQ-RAP & 0.051 & 1378.3 & \textcolor{red}{56.9} & \textcolor{red}{0.79} & 0.2937 & 126.86 & 0.6827 \\
        VARSR & 0.267 & 12341.0 & 1101.9 & 8.73 & 0.3308 & 135.52 & \textcolor{red}{0.7057} \\
        Ours & \textcolor{blue}{0.047} & \textcolor{blue}{572.1} & \textcolor{blue}{339.0} & \textcolor{blue}{1.88} & \textcolor{red}{0.2477} & \textcolor{red}{106.43} & \textcolor{blue}{0.6924} \\
        \bottomrule
    \end{tabular}
    }
    \label{tab:efficiency_comp}
\end{table}

\paragraph{Efficiency Comparisons.}
We further compare the inference efficiency of HiTokSR against representative diffusion-based (OSEDiff, AdcSR) and VQ-based (TVQ-RAP, VARSR) methods. \Cref{tab:efficiency_comp} reports runtime, MACs, parameter count, GPU memory, and perceptual quality metrics (LPIPS, FID, CLIPIQA) measured on the RealSR dataset with $128\times128$ inputs on a single NVIDIA 4090 GPU.
HiTokSR achieves an inference time of 0.047 seconds, on par with the fastest method AdcSR, while attaining the best LPIPS and FID. Its CLIPIQA score approaches that of VARSR, despite the latter requiring over $20\times$ more MACs.
% With a moderate parameter count and GPU memory footprint, HiTokSR offers a favorable efficiency-quality trade-off: it combines the speed of the lightest single-pass frameworks with perceptual quality competitive against much heavier alternatives.

\subsection{Ablation Study}
\paragraph{Tokenizer Designs.}
% \begin{table}[t]
%     \centering
%     \caption{Ablation on tokenizer designs.}
%     \resizebox{0.8\textwidth}{!}{%
%     \begin{tabular}{c | c c c | c c c}
%         \toprule
%         Methods & r-PSNR& r-LPIPS & r-FID & g-PSNR & g-LPIPS & g-FID
%         \\
%         \midrule
%         Vanilla VQ (baseline) & 23.90 & 0.1986 & 34.74 & 24.43 & 0.3132 & 176.02
%         \\
%         + Multi-Group Codebook & 25.49 & 0.1114 & 18.27 & 25.01 & 0.2907 & 156.74
%         \\
%         + Group-Masking & 26.31 & 0.1068 & 15.81 & 25.31 & 0.2833 & 140.51
%         \\
%         + Frequency-Matched Supervision & 26.74 & 0.0677 & 12.05 & 25.63 & 0.2628 & 132.63
%         \\
%         \bottomrule
%     \end{tabular}
%     }
%     \label{tab:abtest_tok_design}
% \end{table}
\Cref{tab:abtest_tok_design} reports reconstruction and generation metrics on DIV2K and RealLR under different tokenizer designs. All variants are trained for 500k iterations, and the generator is subsequently trained for 300k iterations with cross-entropy loss only.
Compared to vanilla VQ, multi-group quantization partitions the latent space into lower-dimensional subspaces, substantially improving both reconstruction and generation quality by mitigating the curse of dimensionality.
Building on this, the group masking strategy implicitly encourages coarse-to-fine spectral decomposition: stochastically masking later groups forces early groups to capture coarse structures while later groups learn residual details, yielding consistent gains across all metrics.
\begin{wraptable}{r}{0.7\textwidth}  % r = 靠右，宽度为表格实际占用宽度
    \centering
    \caption{Ablation on tokenizer designs.}
    \label{tab:abtest_tok_design}
    \resizebox{0.7\textwidth}{!}{%
    \begin{tabular}{c | c c c | c c c}
        \toprule
        Methods & r-PSNR & r-LPIPS & r-FID & g-PSNR & g-LPIPS & g-FID \\
        \midrule
        Vanilla VQ (baseline) & 23.90 & 0.1986 & 34.74 & 24.43 & 0.3132 & 176.02 \\
        + Multi-Group Codebook & 25.49 & 0.1114 & 18.27 & 25.01 & 0.2907 & 156.74 \\
        + Group-Masking & 26.31 & 0.1068 & 15.81 & 25.31 & 0.2833 & 140.51 \\
        + Frequency-Matched Supervision & 26.74 & 0.0677 & 12.05 & 25.63 & 0.2628 & 132.63 \\
        \bottomrule
    \end{tabular}
    }
\end{wraptable}
Finally, we enforce an explicit mapping between group indices and frequency bands by pairing the number of active groups with DWT decomposition levels, which achieves the best results across all metrics.
These results confirm that the three components are complementary: grouping enables stable masking, masking creates the condition for frequency specialization, and explicit supervision provides the necessary guidance for a clean structure--texture separation.

\paragraph{Semantic Guidance.}
\begin{wraptable}{r}{0.5\textwidth}
    \centering
    \caption{Ablation on design choices for incorporating VFM semantic priors.  All results are obtained under the same generator training setting, \textit{without} decoder fine-tuning.}
    \resizebox{0.5\textwidth}{!}{%
    \begin{tabular}{c | c c c c | c c}
        \toprule
        Exp. & SFT & CA & CLS  & MS-CLS & LPIPS & CLIPIQA
        \\
        \midrule
        (a) & & & & & 0.2735 & 0.6327
        \\
        (b) & \checkmark & & & & 0.2503 & 0.6564
        \\
        (c) & & \checkmark & & & 0.2531 & 0.6651
        \\
        (d) & \checkmark & & \checkmark & & 0.2513 & 0.6801
        \\
        (e) & \checkmark & & & \checkmark & 0.2442 & 0.6747
        \\
        \bottomrule
    \end{tabular}
    }
    \label{tab:abtest_guidance_design}
\end{wraptable}
We investigate several design choices for incorporating semantic priors in \Cref{tab:abtest_guidance_design}.
The comparison between Exps.~(a), (b), and (c) confirms the effectiveness of semantic guidance.
While cross-attention (Exp.~(c)) yields higher CLIPIQA than SFT (Exp.~(b)), it achieves similar LPIPS and incurs higher inference overhead, making SFT a more practical choice.
Adding a single global CLS token on top of SFT (Exp.~(d)) further improves CLIPIQA, as explicit scene-level context complements spatially dense modulation.
Our full design (Exp.~(e)) replaces the single CLS token with multi-scale CLS tokens, achieving the best LPIPS with a slight CLIPIQA drop, which suggests that multi-scale tokens achieve a better perception–fidelity balance.

\paragraph{Decoder Finetuning.}
\Cref{tab:abtest_decoder_topk} compares three decoder configurations.
\textit{Frozen} keeps the decoder fixed after generator training, leaving the train--test token discrepancy unaddressed. 
\begin{wraptable}{r}{0.4\textwidth}
    \centering
    \caption{Ablation on decoder fine-tuning strategies.}
    \resizebox{0.4\textwidth}{!}{%
    \begin{tabular}{c | c c c c}
        \toprule
        Methods & PSNR & LPIPS & FID & CLIPIQA
        \\
        \midrule
        Frozen & 26.04 & 0.2442 & 110.97 & 0.6747
        \\
        Top-1 & 26.48 & 0.2517 & 115.53 & 0.6586
        \\
        Top-5 & 26.16 & 0.2477 & 106.43 & 0.6924
        \\
        Top-20 & 26.18 & 0.2473 & 107.97 & 0.6892
        \\
        \bottomrule
    \end{tabular}
    }
    \label{tab:abtest_decoder_topk}
\end{wraptable}
It attains reasonable LPIPS and CLIPIQA but trails in FID, indicating that the discrepancy primarily degrades distribution-level realism rather than per-sample perceptual quality.
\textit{Top-1} fine-tunes the decoder using the generator's top-1 prediction. Although this exposes the decoder to prediction errors, the errors follow a fixed bias, causing the decoder to overfit to the generator.
\textit{Top-5} samples from the top-5 predictions during fine-tuning, introducing stochastic perturbation that exposes the decoder to diverse token errors. This prevents overfitting to any specific error pattern and yields robustness to the range of mistakes the generator makes at inference. 
\textit{Top-20} further expands the sampling range and achieves similarly strong results, suggesting that a moderate degree of perturbation (Top-5) already suffices to realize the benefits of decoder fine-tuning.

\section{Conclusion}
\label{sec:conclusion}

In this paper, we present HiTokSR, a hierarchical token prediction framework that addresses the limitations of monolithic vector quantization in real-world image super-resolution. Conventional single-codebook methods struggle with the entanglement of low-frequency structures and high-frequency textures, creating a persistent trade-off between representational capacity and optimization stability. To resolve this, HiTokSR partitions the latent space into channel-wise groups with independent sub-codebooks, employing a group masking strategy for implicit spectral decomposition and frequency-matched DWT supervision for explicit structure–texture disentanglement. The full framework further integrates semantic priors from a vision foundation model and a decoder fine-tuning stage to bridge the train–test token discrepancy. Evaluations on real-world benchmarks demonstrate that HiTokSR achieves an effective balance among inference efficiency, perceptual quality, and reconstruction fidelity. We hope our frequency-aware tokenization design can provide practical insights and inspire future research in VQ-based image generation and restoration.

% This paper presented HiTokSR, a hierarchical token prediction framework for real-world image super-resolution that replaces monolithic vector quantization with a coarse-to-fine, frequency-aware tokenizer.
% Conventional VQ-based SR methods rely on monolithic codebooks that entangle low-frequency structures and high-frequency textures, creating a trade-off between representational capacity and optimization stability.
% To address this, HiTokSR partitions the latent space into channel-wise groups with independent sub-codebooks, uses a group masking strategy to drive implicit spectral decomposition, and applies frequency-matched DWT supervision for explicit structure--texture disentanglement.
% The full framework further incorporates semantic priors from a vision foundation model and a decoder fine-tuning stage to bridge the train-test token discrepancy.
% On real-world benchmarks, HiTokSR achieves an effective balance among inference efficiency, perceptual quality, and reconstruction fidelity, establishing a practical coarse-to-fine tokenization framework for real-world super-resolution.
% We hope the frequency-aware tokenization design could inspire future work in VQ-based image generation and restoration.

% \section*{References}
% {
%     \small
%     \bibliographystyle{acl_natbib}
%     \bibliography{reference}
% }

{
\small
\bibliographystyle{acl_natbib}
\bibliography{reference}

@article{NIQE,
  author       = {Anish Mittal and
                  Rajiv Soundararajan and
                  Alan C. Bovik},
  title        = {Making a "Completely Blind" Image Quality Analyzer},
  journal      = {{IEEE} Signal Process. Lett.},
  volume       = {20},
  number       = {3},
  pages        = {209--212},
  year         = {2013},
  url          = {https://doi.org/10.1109/LSP.2012.2227726},
  doi          = {10.1109/LSP.2012.2227726},
  bibsource    = {dblp computer science bibliography, https://dblp.org}
}

@article{DISTS,
  author       = {Keyan Ding and
                  Kede Ma and
                  Shiqi Wang and
                  Eero P. Simoncelli},
  title        = {Image Quality Assessment: Unifying Structure and Texture Similarity},
  journal      = {CoRR},
  volume       = {abs/2004.07728},
  year         = {2020},
  url          = {https://arxiv.org/abs/2004.07728},
  eprinttype    = {arXiv},
  eprint       = {2004.07728},
  bibsource    = {dblp computer science bibliography, https://dblp.org}
}

@inproceedings{ESRGAN,
  author    = {Xintao Wang and
               Ke Yu and
               Shixiang Wu and
               Jinjin Gu and
               Yihao Liu and
               Chao Dong and
               Yu Qiao and
               Chen Change Loy},
  editor    = {Laura Leal{-}Taix{\'{e}} and
               Stefan Roth},
  title     = {{ESRGAN:} Enhanced Super-Resolution Generative Adversarial Networks},
  booktitle = {Computer Vision - {ECCV} 2018 Workshops - Munich, Germany, September
               8-14, 2018, Proceedings, Part {V}},
  series    = {Lecture Notes in Computer Science},
  volume    = {11133},
  pages     = {63--79},
  publisher = {Springer},
  year      = {2018},
  url       = {https://doi.org/10.1007/978-3-030-11021-5\_5},
  doi       = {10.1007/978-3-030-11021-5\_5},
  bibsource = {dblp computer science bibliography, https://dblp.org}
}

@inproceedings{SRGAN,
  author    = {Christian Ledig and
               Lucas Theis and
               Ferenc Huszar and
               Jose Caballero and
               Andrew Cunningham and
               Alejandro Acosta and
               Andrew P. Aitken and
               Alykhan Tejani and
               Johannes Totz and
               Zehan Wang and
               Wenzhe Shi},
  title     = {Photo-Realistic Single Image Super-Resolution Using a Generative Adversarial
               Network},
  booktitle = {2017 {IEEE} Conference on Computer Vision and Pattern Recognition,
               {CVPR} 2017, Honolulu, HI, USA, July 21-26, 2017},
  pages     = {105--114},
  publisher = {{IEEE} Computer Society},
  year      = {2017},
  url       = {https://doi.org/10.1109/CVPR.2017.19},
  doi       = {10.1109/CVPR.2017.19},
  bibsource = {dblp computer science bibliography, https://dblp.org}
}

@inproceedings{LDL,
  title={Details or Artifacts: A Locally Discriminative Learning Approach to Realistic Image Super-Resolution},
  author={Liang, Jie and Zeng, Hui and Zhang, Lei},
  booktitle={Proceedings of the IEEE Conference on Computer Vision and Pattern Recognition},
  year={2022}
}

@inproceedings{SPSR,
  title={Structure-Preserving Super Resolution with Gradient Guidance},
  author={Ma, Cheng and Rao, Yongming and Cheng, Yean and Chen, Ce and Lu, Jiwen and Zhou, Jie},
  booktitle={Proceedings of the IEEE Conference on Computer Vision and Pattern Recognition (CVPR)},
  year={2020}
}

@inproceedings{DualFormer,
	title={On the Effectiveness of Spectral Discriminators for Perceptual Quality Improvement},
	author={Luo, Xin and Zhu, Yunan and Xu, Shunxin and Liu, Dong},
	booktitle={ICCV},
	year={2023}
}

@inproceedings{DRealSR,
  author       = {Pengxu Wei and
                  Ziwei Xie and
                  Hannan Lu and
                  Zongyuan Zhan and
                  Qixiang Ye and
                  Wangmeng Zuo and
                  Liang Lin},
  editor       = {Andrea Vedaldi and
                  Horst Bischof and
                  Thomas Brox and
                  Jan{-}Michael Frahm},
  title        = {Component Divide-and-Conquer for Real-World Image Super-Resolution},
  booktitle    = {Computer Vision - {ECCV} 2020 - 16th European Conference, Glasgow,
                  UK, August 23-28, 2020, Proceedings, Part {VIII}},
  series       = {Lecture Notes in Computer Science},
  volume       = {12353},
  pages        = {101--117},
  publisher    = {Springer},
  year         = {2020},
  url          = {https://doi.org/10.1007/978-3-030-58598-3\_7},
  doi          = {10.1007/978-3-030-58598-3\_7},
  bibsource    = {dblp computer science bibliography, https://dblp.org}
}

@inproceedings{LPIPS,
  author    = {Richard Zhang and
               Phillip Isola and
               Alexei A. Efros and
               Eli Shechtman and
               Oliver Wang},
  title     = {The Unreasonable Effectiveness of Deep Features as a Perceptual Metric},
  booktitle = {2018 {IEEE} Conference on Computer Vision and Pattern Recognition,
               {CVPR} 2018, Salt Lake City, UT, USA, June 18-22, 2018},
  pages     = {586--595},
  publisher = {Computer Vision Foundation / {IEEE} Computer Society},
  year      = {2018},
  url       = {http://openaccess.thecvf.com/content\_cvpr\_2018/html/Zhang\_The\_Unreasonable\_Effectiveness\_CVPR\_2018\_paper.html},
  doi       = {10.1109/CVPR.2018.00068},
  bibsource = {dblp computer science bibliography, https://dblp.org}
}

@inproceedings{realesrgan,
  author       = {Xintao Wang and
                  Liangbin Xie and
                  Chao Dong and
                  Ying Shan},
  title        = {Real-ESRGAN: Training Real-World Blind Super-Resolution with Pure
                  Synthetic Data},
  booktitle    = {{IEEE/CVF} International Conference on Computer Vision Workshops,
                  {ICCVW} 2021, Montreal, QC, Canada, October 11-17, 2021},
  pages        = {1905--1914},
  publisher    = {{IEEE}},
  year         = {2021},
  url          = {https://doi.org/10.1109/ICCVW54120.2021.00217},
  doi          = {10.1109/ICCVW54120.2021.00217},
  bibsource    = {dblp computer science bibliography, https://dblp.org}
}

@article{BSRGAN,
  author    = {Kai Zhang and
               Jingyun Liang and
               Luc Van Gool and
               Radu Timofte},
  title     = {Designing a Practical Degradation Model for Deep Blind Image Super-Resolution},
  journal   = {CoRR},
  volume    = {abs/2103.14006},
  year      = {2021},
  url       = {https://arxiv.org/abs/2103.14006},
  eprinttype = {arXiv},
  eprint    = {2103.14006},
  bibsource = {dblp computer science bibliography, https://dblp.org}
}

@inproceedings{FeMaSR,
  author       = {Chaofeng Chen and
                  Xinyu Shi and
                  Yipeng Qin and
                  Xiaoming Li and
                  Xiaoguang Han and
                  Tao Yang and
                  Shihui Guo},
  editor       = {Jo{\~{a}}o Magalh{\~{a}}es and
                  Alberto Del Bimbo and
                  Shin'ichi Satoh and
                  Nicu Sebe and
                  Xavier Alameda{-}Pineda and
                  Qin Jin and
                  Vincent Oria and
                  Laura Toni},
  title        = {Real-World Blind Super-Resolution via Feature Matching with Implicit
                  High-Resolution Priors},
  booktitle    = {{MM} '22: The 30th {ACM} International Conference on Multimedia, Lisboa,
                  Portugal, October 10 - 14, 2022},
  pages        = {1329--1338},
  publisher    = {{ACM}},
  year         = {2022},
  url          = {https://doi.org/10.1145/3503161.3547833},
  doi          = {10.1145/3503161.3547833},
  bibsource    = {dblp computer science bibliography, https://dblp.org}
}

@inproceedings{OSEDiff,
  author       = {Rongyuan Wu and
                  Lingchen Sun and
                  Zhiyuan Ma and
                  Lei Zhang},
  editor       = {Amir Globersons and
                  Lester Mackey and
                  Danielle Belgrave and
                  Angela Fan and
                  Ulrich Paquet and
                  Jakub M. Tomczak and
                  Cheng Zhang},
  title        = {One-Step Effective Diffusion Network for Real-World Image Super-Resolution},
  booktitle    = {Advances in Neural Information Processing Systems 38: Annual Conference
                  on Neural Information Processing Systems 2024, NeurIPS 2024, Vancouver,
                  BC, Canada, December 10 - 15, 2024},
  year         = {2024},
  url          = {http://papers.nips.cc/paper\_files/paper/2024/hash/a8223b0ad64007423ffb308b0dd92298-Abstract-Conference.html},
  bibsource    = {dblp computer science bibliography, https://dblp.org}
}

@inproceedings{VARSR,
  author       = {Yunpeng Qu and
                  Kun Yuan and
                  Jinhua Hao and
                  Kai Zhao and
                  Qizhi Xie and
                  Ming Sun and
                  Chao Zhou},
  editor       = {Aarti Singh and
                  Maryam Fazel and
                  Daniel Hsu and
                  Simon Lacoste{-}Julien and
                  Felix Berkenkamp and
                  Tegan Maharaj and
                  Kiri Wagstaff and
                  Jerry Zhu},
  title        = {Visual Autoregressive Modeling for Image Super-Resolution},
  booktitle    = {Forty-second International Conference on Machine Learning, {ICML}
                  2025, Vancouver, BC, Canada, July 13-19, 2025},
  series       = {Proceedings of Machine Learning Research},
  publisher    = {{PMLR} / OpenReview.net},
  year         = {2025},
  url          = {https://proceedings.mlr.press/v267/qu25h.html},
  bibsource    = {dblp computer science bibliography, https://dblp.org}
}

@inproceedings{VAR,
  author       = {Keyu Tian and
                  Yi Jiang and
                  Zehuan Yuan and
                  Bingyue Peng and
                  Liwei Wang},
  editor       = {Amir Globersons and
                  Lester Mackey and
                  Danielle Belgrave and
                  Angela Fan and
                  Ulrich Paquet and
                  Jakub M. Tomczak and
                  Cheng Zhang},
  title        = {Visual Autoregressive Modeling: Scalable Image Generation via Next-Scale
                  Prediction},
  booktitle    = {Advances in Neural Information Processing Systems 38: Annual Conference
                  on Neural Information Processing Systems 2024, NeurIPS 2024, Vancouver,
                  BC, Canada, December 10 - 15, 2024},
  year         = {2024},
  url          = {http://papers.nips.cc/paper\_files/paper/2024/hash/9a24e284b187f662681440ba15c416fb-Abstract-Conference.html},
  bibsource    = {dblp computer science bibliography, https://dblp.org}
}

@inproceedings{Codeformer,
  author       = {Shangchen Zhou and
                  Kelvin C. K. Chan and
                  Chongyi Li and
                  Chen Change Loy},
  editor       = {Sanmi Koyejo and
                  S. Mohamed and
                  A. Agarwal and
                  Danielle Belgrave and
                  K. Cho and
                  A. Oh},
  title        = {Towards Robust Blind Face Restoration with Codebook Lookup Transformer},
  booktitle    = {Advances in Neural Information Processing Systems 35: Annual Conference
                  on Neural Information Processing Systems 2022, NeurIPS 2022, New Orleans,
                  LA, USA, November 28 - December 9, 2022},
  year         = {2022},
  url          = {http://papers.nips.cc/paper\_files/paper/2022/hash/c573258c38d0a3919d8c1364053c45df-Abstract-Conference.html},
  bibsource    = {dblp computer science bibliography, https://dblp.org}
}

@article{TVQ,
  author       = {Qifan Li and
                  Jiale Zou and
                  Jinhua Zhang and
                  Wei Long and
                  Xingyu Zhou and
                  Shuhang Gu},
  title        = {Texture Vector-Quantization and Reconstruction Aware Prediction for
                  Generative Super-Resolution},
  journal      = {CoRR},
  volume       = {abs/2509.23774},
  year         = {2025},
  url          = {https://doi.org/10.48550/arXiv.2509.23774},
  doi          = {10.48550/ARXIV.2509.23774},
  eprinttype   = {arXiv},
  eprint       = {2509.23774},
  bibsource    = {dblp computer science bibliography, https://dblp.org}
}

@article{StableSR,
  author       = {Jianyi Wang and
                  Zongsheng Yue and
                  Shangchen Zhou and
                  Kelvin C. K. Chan and
                  Chen Change Loy},
  title        = {Exploiting Diffusion Prior for Real-World Image Super-Resolution},
  journal      = {Int. J. Comput. Vis.},
  volume       = {132},
  number       = {12},
  pages        = {5929--5949},
  year         = {2024},
  url          = {https://doi.org/10.1007/s11263-024-02168-7},
  doi          = {10.1007/S11263-024-02168-7},
  bibsource    = {dblp computer science bibliography, https://dblp.org}
}

@article{DiffBIR,
  author       = {Xinqi Lin and
                  Jingwen He and
                  Ziyan Chen and
                  Zhaoyang Lyu and
                  Ben Fei and
                  Bo Dai and
                  Wanli Ouyang and
                  Yu Qiao and
                  Chao Dong},
  title        = {DiffBIR: Towards Blind Image Restoration with Generative Diffusion
                  Prior},
  journal      = {CoRR},
  volume       = {abs/2308.15070},
  year         = {2023},
  url          = {https://doi.org/10.48550/arXiv.2308.15070},
  doi          = {10.48550/ARXIV.2308.15070},
  eprinttype   = {arXiv},
  eprint       = {2308.15070},
  bibsource    = {dblp computer science bibliography, https://dblp.org}
}

@inproceedings{SeeSR,
  author       = {Rongyuan Wu and
                  Tao Yang and
                  Lingchen Sun and
                  Zhengqiang Zhang and
                  Shuai Li and
                  Lei Zhang},
  title        = {SeeSR: Towards Semantics-Aware Real-World Image Super-Resolution},
  booktitle    = {{IEEE/CVF} Conference on Computer Vision and Pattern Recognition,
                  {CVPR} 2024, Seattle, WA, USA, June 16-22, 2024},
  pages        = {25456--25467},
  publisher    = {{IEEE}},
  year         = {2024},
  url          = {https://doi.org/10.1109/CVPR52733.2024.02405},
  doi          = {10.1109/CVPR52733.2024.02405},
  bibsource    = {dblp computer science bibliography, https://dblp.org}
}

@inproceedings{Resshift,
  author       = {Zongsheng Yue and
                  Jianyi Wang and
                  Chen Change Loy},
  editor       = {Alice Oh and
                  Tristan Naumann and
                  Amir Globerson and
                  Kate Saenko and
                  Moritz Hardt and
                  Sergey Levine},
  title        = {ResShift: Efficient Diffusion Model for Image Super-resolution by
                  Residual Shifting},
  booktitle    = {Advances in Neural Information Processing Systems 36: Annual Conference
                  on Neural Information Processing Systems 2023, NeurIPS 2023, New Orleans,
                  LA, USA, December 10 - 16, 2023},
  year         = {2023},
  url          = {http://papers.nips.cc/paper\_files/paper/2023/hash/2ac2eac5098dba08208807b65c5851cc-Abstract-Conference.html},
  bibsource    = {dblp computer science bibliography, https://dblp.org}
}

@inproceedings{SinSR,
  author       = {Yufei Wang and
                  Wenhan Yang and
                  Xinyuan Chen and
                  Yaohui Wang and
                  Lanqing Guo and
                  Lap{-}Pui Chau and
                  Ziwei Liu and
                  Yu Qiao and
                  Alex C. Kot and
                  Bihan Wen},
  title        = {SinSR: Diffusion-Based Image Super-Resolution in a Single Step},
  booktitle    = {{IEEE/CVF} Conference on Computer Vision and Pattern Recognition,
                  {CVPR} 2024, Seattle, WA, USA, June 16-22, 2024},
  pages        = {25796--25805},
  publisher    = {{IEEE}},
  year         = {2024},
  url          = {https://doi.org/10.1109/CVPR52733.2024.02437},
  doi          = {10.1109/CVPR52733.2024.02437},
  bibsource    = {dblp computer science bibliography, https://dblp.org}
}

@inproceedings{adcsr,
  author       = {Bin Chen and
                  Gehui Li and
                  Rongyuan Wu and
                  Xindong Zhang and
                  Jie Chen and
                  Jian Zhang and
                  Lei Zhang},
  title        = {Adversarial Diffusion Compression for Real-World Image Super-Resolution},
  booktitle    = {{IEEE/CVF} Conference on Computer Vision and Pattern Recognition,
                  {CVPR} 2025, Nashville, TN, USA, June 11-15, 2025},
  pages        = {28208--28220},
  publisher    = {Computer Vision Foundation / {IEEE}},
  year         = {2025},
  url          = {https://openaccess.thecvf.com/content/CVPR2025/html/Chen\_Adversarial\_Diffusion\_Compression\_for\_Real-World\_Image\_Super-Resolution\_CVPR\_2025\_paper.html},
  doi          = {10.1109/CVPR52734.2025.02627},
  bibsource    = {dblp computer science bibliography, https://dblp.org}
}

@inproceedings{TSDSR,
  author       = {Linwei Dong and
                  Qingnan Fan and
                  Yihong Guo and
                  Zhonghao Wang and
                  Qi Zhang and
                  Jinwei Chen and
                  Yawei Luo and
                  Changqing Zou},
  title        = {{TSD-SR:} One-Step Diffusion with Target Score Distillation for Real-World
                  Image Super-Resolution},
  booktitle    = {{IEEE/CVF} Conference on Computer Vision and Pattern Recognition,
                  {CVPR} 2025, Nashville, TN, USA, June 11-15, 2025},
  pages        = {23174--23184},
  publisher    = {Computer Vision Foundation / {IEEE}},
  year         = {2025},
  url          = {https://openaccess.thecvf.com/content/CVPR2025/html/Dong\_TSD-SR\_One-Step\_Diffusion\_with\_Target\_Score\_Distillation\_for\_Real-World\_Image\_CVPR\_2025\_paper.html},
  doi          = {10.1109/CVPR52734.2025.02158},
  bibsource    = {dblp computer science bibliography, https://dblp.org}
}

@article{TinySR,
  author       = {Linwei Dong and
                  Qingnan Fan and
                  Yuhang Yu and
                  Qi Zhang and
                  Jinwei Chen and
                  Yawei Luo and
                  Changqing Zou},
  title        = {TinySR: Pruning Diffusion for Real-World Image Super-Resolution},
  journal      = {CoRR},
  volume       = {abs/2508.17434},
  year         = {2025},
  url          = {https://doi.org/10.48550/arXiv.2508.17434},
  doi          = {10.48550/ARXIV.2508.17434},
  eprinttype   = {arXiv},
  eprint       = {2508.17434},
  bibsource    = {dblp computer science bibliography, https://dblp.org}
}

@inproceedings{FaithDiff,
  author       = {Junyang Chen and
                  Jinshan Pan and
                  Jiangxin Dong},
  title        = {FaithDiff: Unleashing Diffusion Priors for Faithful Image Super-resolution},
  booktitle    = {{IEEE/CVF} Conference on Computer Vision and Pattern Recognition,
                  {CVPR} 2025, Nashville, TN, USA, June 11-15, 2025},
  pages        = {28188--28197},
  publisher    = {Computer Vision Foundation / {IEEE}},
  year         = {2025},
  url          = {https://openaccess.thecvf.com/content/CVPR2025/html/Chen\_FaithDiff\_Unleashing\_Diffusion\_Priors\_for\_Faithful\_Image\_Super-resolution\_CVPR\_2025\_paper.html},
  doi          = {10.1109/CVPR52734.2025.02625},
  bibsource    = {dblp computer science bibliography, https://dblp.org}
}

@inproceedings{pisasr,
  author       = {Lingchen Sun and
                  Rongyuan Wu and
                  Zhiyuan Ma and
                  Shuaizheng Liu and
                  Qiaosi Yi and
                  Lei Zhang},
  title        = {Pixel-level and Semantic-level Adjustable Super-resolution: {A} Dual-LoRA
                  Approach},
  booktitle    = {{IEEE/CVF} Conference on Computer Vision and Pattern Recognition,
                  {CVPR} 2025, Nashville, TN, USA, June 11-15, 2025},
  pages        = {2333--2343},
  publisher    = {Computer Vision Foundation / {IEEE}},
  year         = {2025},
  url          = {https://openaccess.thecvf.com/content/CVPR2025/html/Sun\_Pixel-level\_and\_Semantic-level\_Adjustable\_Super-resolution\_A\_Dual-LoRA\_Approach\_CVPR\_2025\_paper.html},
  doi          = {10.1109/CVPR52734.2025.00223},
  bibsource    = {dblp computer science bibliography, https://dblp.org}
}

@inproceedings{vqgan,
  author       = {Patrick Esser and
                  Robin Rombach and
                  Bj{\"{o}}rn Ommer},
  title        = {Taming Transformers for High-Resolution Image Synthesis},
  booktitle    = {{IEEE} Conference on Computer Vision and Pattern Recognition, {CVPR}
                  2021, virtual, June 19-25, 2021},
  pages        = {12873--12883},
  publisher    = {Computer Vision Foundation / {IEEE}},
  year         = {2021},
  url          = {https://openaccess.thecvf.com/content/CVPR2021/html/Esser\_Taming\_Transformers\_for\_High-Resolution\_Image\_Synthesis\_CVPR\_2021\_paper.html},
  doi          = {10.1109/CVPR46437.2021.01268},
  bibsource    = {dblp computer science bibliography, https://dblp.org}
}

@inproceedings{vqvae,
  author       = {A{\"{a}}ron van den Oord and
                  Oriol Vinyals and
                  Koray Kavukcuoglu},
  editor       = {Isabelle Guyon and
                  Ulrike von Luxburg and
                  Samy Bengio and
                  Hanna M. Wallach and
                  Rob Fergus and
                  S. V. N. Vishwanathan and
                  Roman Garnett},
  title        = {Neural Discrete Representation Learning},
  booktitle    = {Advances in Neural Information Processing Systems 30: Annual Conference
                  on Neural Information Processing Systems 2017, December 4-9, 2017,
                  Long Beach, CA, {USA}},
  pages        = {6306--6315},
  year         = {2017},
  url          = {https://proceedings.neurips.cc/paper/2017/hash/7a98af17e63a0ac09ce2e96d03992fbc-Abstract.html},
  bibsource    = {dblp computer science bibliography, https://dblp.org}
}

@article{vqgan-lc,
  author       = {Lei Zhu and
                  Fangyun Wei and
                  Yanye Lu and
                  Dong Chen},
  title        = {Scaling the Codebook Size of {VQGAN} to 100,000 with a Utilization
                  Rate of 99{\%}},
  journal      = {CoRR},
  volume       = {abs/2406.11837},
  year         = {2024},
  url          = {https://doi.org/10.48550/arXiv.2406.11837},
  doi          = {10.48550/ARXIV.2406.11837},
  eprinttype   = {arXiv},
  eprint       = {2406.11837},
  bibsource    = {dblp computer science bibliography, https://dblp.org}
}

@inproceedings{rqvae,
  author       = {Doyup Lee and
                  Chiheon Kim and
                  Saehoon Kim and
                  Minsu Cho and
                  Wook{-}Shin Han},
  title        = {Autoregressive Image Generation using Residual Quantization},
  booktitle    = {{IEEE/CVF} Conference on Computer Vision and Pattern Recognition,
                  {CVPR} 2022, New Orleans, LA, USA, June 18-24, 2022},
  pages        = {11513--11522},
  publisher    = {{IEEE}},
  year         = {2022},
  url          = {https://doi.org/10.1109/CVPR52688.2022.01123},
  doi          = {10.1109/CVPR52688.2022.01123},
  bibsource    = {dblp computer science bibliography, https://dblp.org}
}

@inproceedings{Dreamclear,
  author       = {Yuang Ai and
                  Xiaoqiang Zhou and
                  Huaibo Huang and
                  Xiaotian Han and
                  Zhengyu Chen and
                  Quanzeng You and
                  Hongxia Yang},
  editor       = {Amir Globersons and
                  Lester Mackey and
                  Danielle Belgrave and
                  Angela Fan and
                  Ulrich Paquet and
                  Jakub M. Tomczak and
                  Cheng Zhang},
  title        = {DreamClear: High-Capacity Real-World Image Restoration with Privacy-Safe
                  Dataset Curation},
  booktitle    = {Advances in Neural Information Processing Systems 38: Annual Conference
                  on Neural Information Processing Systems 2024, NeurIPS 2024, Vancouver,
                  BC, Canada, December 10 - 15, 2024},
  year         = {2024},
  url          = {http://papers.nips.cc/paper\_files/paper/2024/hash/6452474601429509f3035dc81c233226-Abstract-Conference.html},
  bibsource    = {dblp computer science bibliography, https://dblp.org}
}

@inproceedings{RealSR,
  author       = {Jianrui Cai and
                  Hui Zeng and
                  Hongwei Yong and
                  Zisheng Cao and
                  Lei Zhang},
  title        = {Toward Real-World Single Image Super-Resolution: {A} New Benchmark
                  and a New Model},
  booktitle    = {2019 {IEEE/CVF} International Conference on Computer Vision, {ICCV}
                  2019, Seoul, Korea (South), October 27 - November 2, 2019},
  pages        = {3086--3095},
  publisher    = {{IEEE}},
  year         = {2019},
  url          = {https://doi.org/10.1109/ICCV.2019.00318},
  doi          = {10.1109/ICCV.2019.00318},
  bibsource    = {dblp computer science bibliography, https://dblp.org}
}

@article{Siglip2,
  author       = {Michael Tschannen and
                  Alexey A. Gritsenko and
                  Xiao Wang and
                  Muhammad Ferjad Naeem and
                  Ibrahim Alabdulmohsin and
                  Nikhil Parthasarathy and
                  Talfan Evans and
                  Lucas Beyer and
                  Ye Xia and
                  Basil Mustafa and
                  Olivier J. H{\'{e}}naff and
                  Jeremiah Harmsen and
                  Andreas Steiner and
                  Xiaohua Zhai},
  title        = {SigLIP 2: Multilingual Vision-Language Encoders with Improved Semantic
                  Understanding, Localization, and Dense Features},
  journal      = {CoRR},
  volume       = {abs/2502.14786},
  year         = {2025},
  url          = {https://doi.org/10.48550/arXiv.2502.14786},
  doi          = {10.48550/ARXIV.2502.14786},
  eprinttype   = {arXiv},
  eprint       = {2502.14786},
  bibsource    = {dblp computer science bibliography, https://dblp.org}
}

@article{dinov3,
  author       = {Oriane Sim{\'{e}}oni and
                  Huy V. Vo and
                  Maximilian Seitzer and
                  Federico Baldassarre and
                  Maxime Oquab and
                  Cijo Jose and
                  Vasil Khalidov and
                  Marc Szafraniec and
                  Seung Eun Yi and
                  Micha{\"{e}}l Ramamonjisoa and
                  Francisco Massa and
                  Daniel Haziza and
                  Luca Wehrstedt and
                  Jianyuan Wang and
                  Timoth{\'{e}}e Darcet and
                  Th{\'{e}}o Moutakanni and
                  Leonel Sentana and
                  Claire Roberts and
                  Andrea Vedaldi and
                  Jamie Tolan and
                  John Brandt and
                  Camille Couprie and
                  Julien Mairal and
                  Herv{\'{e}} J{\'{e}}gou and
                  Patrick Labatut and
                  Piotr Bojanowski},
  title        = {DINOv3},
  journal      = {CoRR},
  volume       = {abs/2508.10104},
  year         = {2025},
  url          = {https://doi.org/10.48550/arXiv.2508.10104},
  doi          = {10.48550/ARXIV.2508.10104},
  eprinttype   = {arXiv},
  eprint       = {2508.10104},
  bibsource    = {dblp computer science bibliography, https://dblp.org}
}

@inproceedings{CLIP,
  author       = {Alec Radford and
                  Jong Wook Kim and
                  Chris Hallacy and
                  Aditya Ramesh and
                  Gabriel Goh and
                  Sandhini Agarwal and
                  Girish Sastry and
                  Amanda Askell and
                  Pamela Mishkin and
                  Jack Clark and
                  Gretchen Krueger and
                  Ilya Sutskever},
  editor       = {Marina Meila and
                  Tong Zhang},
  title        = {Learning Transferable Visual Models From Natural Language Supervision},
  booktitle    = {Proceedings of the 38th International Conference on Machine Learning,
                  {ICML} 2021, 18-24 July 2021, Virtual Event},
  series       = {Proceedings of Machine Learning Research},
  pages        = {8748--8763},
  publisher    = {{PMLR}},
  year         = {2021},
  url          = {http://proceedings.mlr.press/v139/radford21a.html},
  bibsource    = {dblp computer science bibliography, https://dblp.org}
}

@inproceedings{Transformer,
  author       = {Ashish Vaswani and
                  Noam Shazeer and
                  Niki Parmar and
                  Jakob Uszkoreit and
                  Llion Jones and
                  Aidan N. Gomez and
                  Lukasz Kaiser and
                  Illia Polosukhin},
  editor       = {Isabelle Guyon and
                  Ulrike von Luxburg and
                  Samy Bengio and
                  Hanna M. Wallach and
                  Rob Fergus and
                  S. V. N. Vishwanathan and
                  Roman Garnett},
  title        = {Attention is All you Need},
  booktitle    = {Advances in Neural Information Processing Systems 30: Annual Conference
                  on Neural Information Processing Systems 2017, December 4-9, 2017,
                  Long Beach, CA, {USA}},
  pages        = {5998--6008},
  year         = {2017},
  url          = {https://proceedings.neurips.cc/paper/2017/hash/3f5ee243547dee91fbd053c1c4a845aa-Abstract.html},
  bibsource    = {dblp computer science bibliography, https://dblp.org}
}

@inproceedings{LSDIR,
  author       = {Yawei Li and
                  Kai Zhang and
                  Jingyun Liang and
                  Jiezhang Cao and
                  Ce Liu and
                  Rui Gong and
                  Yulun Zhang and
                  Hao Tang and
                  Yun Liu and
                  Denis Demandolx and
                  Rakesh Ranjan and
                  Radu Timofte and
                  Luc Van Gool},
  title        = {{LSDIR:} {A} Large Scale Dataset for Image Restoration},
  booktitle    = {{IEEE/CVF} Conference on Computer Vision and Pattern Recognition,
                  {CVPR} 2023 - Workshops, Vancouver, BC, Canada, June 17-24, 2023},
  pages        = {1775--1787},
  publisher    = {{IEEE}},
  year         = {2023},
  url          = {https://doi.org/10.1109/CVPRW59228.2023.00178},
  doi          = {10.1109/CVPRW59228.2023.00178},
  bibsource    = {dblp computer science bibliography, https://dblp.org}
}

@inproceedings{FFHQ,
  author       = {Tero Karras and
                  Samuli Laine and
                  Timo Aila},
  title        = {A Style-Based Generator Architecture for Generative Adversarial Networks},
  booktitle    = {{IEEE} Conference on Computer Vision and Pattern Recognition, {CVPR}
                  2019, Long Beach, CA, USA, June 16-20, 2019},
  pages        = {4401--4410},
  publisher    = {Computer Vision Foundation / {IEEE}},
  year         = {2019},
  url          = {http://openaccess.thecvf.com/content\_CVPR\_2019/html/Karras\_A\_Style-Based\_Generator\_Architecture\_for\_Generative\_Adversarial\_Networks\_CVPR\_2019\_paper.html},
  doi          = {10.1109/CVPR.2019.00453},
  bibsource    = {dblp computer science bibliography, https://dblp.org}
}

@inproceedings{FID,
  author       = {Martin Heusel and
                  Hubert Ramsauer and
                  Thomas Unterthiner and
                  Bernhard Nessler and
                  Sepp Hochreiter},
  editor       = {Isabelle Guyon and
                  Ulrike von Luxburg and
                  Samy Bengio and
                  Hanna M. Wallach and
                  Rob Fergus and
                  S. V. N. Vishwanathan and
                  Roman Garnett},
  title        = {GANs Trained by a Two Time-Scale Update Rule Converge to a Local Nash
                  Equilibrium},
  booktitle    = {Advances in Neural Information Processing Systems 30: Annual Conference
                  on Neural Information Processing Systems 2017, December 4-9, 2017,
                  Long Beach, CA, {USA}},
  pages        = {6626--6637},
  year         = {2017},
  url          = {https://proceedings.neurips.cc/paper/2017/hash/8a1d694707eb0fefe65871369074926d-Abstract.html},
  bibsource    = {dblp computer science bibliography, https://dblp.org}
}

@inproceedings{CLIPIQA,
  author       = {Jianyi Wang and
                  Kelvin C. K. Chan and
                  Chen Change Loy},
  editor       = {Brian Williams and
                  Yiling Chen and
                  Jennifer Neville},
  title        = {Exploring {CLIP} for Assessing the Look and Feel of Images},
  booktitle    = {Thirty-Seventh {AAAI} Conference on Artificial Intelligence, {AAAI}
                  2023, Thirty-Fifth Conference on Innovative Applications of Artificial
                  Intelligence, {IAAI} 2023, Thirteenth Symposium on Educational Advances
                  in Artificial Intelligence, {EAAI} 2023, Washington, DC, USA, February
                  7-14, 2023},
  pages        = {2555--2563},
  publisher    = {{AAAI} Press},
  year         = {2023},
  url          = {https://doi.org/10.1609/aaai.v37i2.25353},
  doi          = {10.1609/AAAI.V37I2.25353},
  bibsource    = {dblp computer science bibliography, https://dblp.org}
}

@inproceedings{MANIQA,
  author       = {Sidi Yang and
                  Tianhe Wu and
                  Shuwei Shi and
                  Shanshan Lao and
                  Yuan Gong and
                  Mingdeng Cao and
                  Jiahao Wang and
                  Yujiu Yang},
  title        = {{MANIQA:} Multi-dimension Attention Network for No-Reference Image
                  Quality Assessment},
  booktitle    = {{IEEE/CVF} Conference on Computer Vision and Pattern Recognition Workshops,
                  {CVPR} Workshops 2022, New Orleans, LA, USA, June 19-20, 2022},
  pages        = {1190--1199},
  publisher    = {{IEEE}},
  year         = {2022},
  url          = {https://doi.org/10.1109/CVPRW56347.2022.00126},
  doi          = {10.1109/CVPRW56347.2022.00126},
  bibsource    = {dblp computer science bibliography, https://dblp.org}
}

@inproceedings{MUSIQ,
  author       = {Junjie Ke and
                  Qifei Wang and
                  Yilin Wang and
                  Peyman Milanfar and
                  Feng Yang},
  title        = {{MUSIQ:} Multi-scale Image Quality Transformer},
  booktitle    = {2021 {IEEE/CVF} International Conference on Computer Vision, {ICCV}
                  2021, Montreal, QC, Canada, October 10-17, 2021},
  pages        = {5128--5137},
  publisher    = {{IEEE}},
  year         = {2021},
  url          = {https://doi.org/10.1109/ICCV48922.2021.00510},
  doi          = {10.1109/ICCV48922.2021.00510},
  bibsource    = {dblp computer science bibliography, https://dblp.org}
}

@article{SSIM,
  author       = {Zhou Wang and
                  Alan C. Bovik and
                  Hamid R. Sheikh and
                  Eero P. Simoncelli},
  title        = {Image quality assessment: from error visibility to structural similarity},
  journal      = {{IEEE} Trans. Image Process.},
  volume       = {13},
  number       = {4},
  pages        = {600--612},
  year         = {2004},
  url          = {https://doi.org/10.1109/TIP.2003.819861},
  doi          = {10.1109/TIP.2003.819861},
  bibsource    = {dblp computer science bibliography, https://dblp.org}
}

@article{QInsight,
  author       = {Weiqi Li and
                  Xuanyu Zhang and
                  Shijie Zhao and
                  Yabin Zhang and
                  Junlin Li and
                  Li Zhang and
                  Jian Zhang},
  title        = {Q-Insight: Understanding Image Quality via Visual Reinforcement Learning},
  journal      = {CoRR},
  volume       = {abs/2503.22679},
  year         = {2025},
  url          = {https://doi.org/10.48550/arXiv.2503.22679},
  doi          = {10.48550/ARXIV.2503.22679},
  eprinttype   = {arXiv},
  eprint       = {2503.22679},
  bibsource    = {dblp computer science bibliography, https://dblp.org}
}

@article{Unipercept,
  author       = {Shuo Cao and
                  Jiayang Li and
                  Xiaohui Li and
                  Yuandong Pu and
                  Kaiwen Zhu and
                  Yuanting Gao and
                  Siqi Luo and
                  Yi Xin and
                  Qi Qin and
                  Yu Zhou and
                  Xiangyu Chen and
                  Wenlong Zhang and
                  Bin Fu and
                  Yu Qiao and
                  Yihao Liu},
  title        = {UniPercept: Towards Unified Perceptual-Level Image Understanding across
                  Aesthetics, Quality, Structure, and Texture},
  journal      = {CoRR},
  volume       = {abs/2512.21675},
  year         = {2025},
  url          = {https://doi.org/10.48550/arXiv.2512.21675},
  doi          = {10.48550/ARXIV.2512.21675},
  eprinttype   = {arXiv},
  eprint       = {2512.21675},
  bibsource    = {dblp computer science bibliography, https://dblp.org}
}

@article{Visualqualityr1,
  author       = {Tianhe Wu and
                  Jian Zou and
                  Jie Liang and
                  Lei Zhang and
                  Kede Ma},
  title        = {VisualQuality-R1: Reasoning-Induced Image Quality Assessment via Reinforcement
                  Learning to Rank},
  journal      = {CoRR},
  volume       = {abs/2505.14460},
  year         = {2025},
  url          = {https://doi.org/10.48550/arXiv.2505.14460},
  doi          = {10.48550/ARXIV.2505.14460},
  eprinttype   = {arXiv},
  eprint       = {2505.14460},
  bibsource    = {dblp computer science bibliography, https://dblp.org}
}
}

\appendix

\section{Implementation details}
\label{sec:supp_implementation_details}
We present the full training configuration of HiTokSR across its three stages 
(Tokenizer, Generator, and Decoder) in ~\Cref{tab:supp_implementation_details}. 
All stages share the same dataset, patch ratio, and codebook settings. 
They differ in network design, optimizer learning rates, and loss composition, 
reflecting their distinct roles.

\begin{table}[!htb]
    \centering
    \caption{Training details of HiTokSR.}
    \begin{tabular}{c | c c c}
        \toprule
        Stage & Tokenizer & Generator & Decoder
        \\
        \midrule
        Datasets & \multicolumn{3}{c}{LSDIR + FFHQ (first 10K)}
        \\
        \midrule
        Train Device & NVIDIA 4090 & NVIDIA 4090 & NVIDIA 4090
        \\
        Patch Size & 512 & 512 & 512
        \\
        Batch Size & 32 & 64 & 64
        \\
        Train Iters & 600k & 600k & 50k
        \\
        \midrule
        Patch Ratio & 16 & 16 & 16
        \\
        Latent Dim & 32 & 32 & 32
        \\
        Code Group & 4 & 4 & 4
        \\
        Code Size & 4096 & 4096 & 4096
        \\
        Code Capacity & $1024^4=2^{40}$ & $1024^4=2^{40}$ & $1024^4=2^{40}$
        \\
        \midrule
        Encoder Depth & 12 & - & 12
        \\
        Encoder Dim & 768 & - & 768
        \\
        Decoder Depth & 12 & 12 & 12
        \\
        Decoder Dim & 768 & 768 & 768
        \\
        Generator Depth & - & 12 & 12
        \\
        Generator Dim & - & 768 & 768
        \\
        \midrule
        Optimizer G & AdamW & AdamW & AdamW
        \\
        Optimizer D & AdamW & AdamW & AdamW
        \\
        Learning Rate G & 1e-4 & 5e-5 & 5e-5
        \\
        Learning Rate D & 5e-5 & 1e-5 & 1e-5
        \\
        \midrule
        L1 Loss Strength & 1.0 & 1.0 & 1.0 
        \\
        Commitment Loss Strength & 0.25 & - & -
        \\
        CrossEntropy Loss Strength & - & 1.0 & -
        \\
        REPA Loss Strength & - & 1.0 & -
        \\
        LPIPS Loss Strength & 1.0 & 1.0 & 1.0
        \\
        Adversarial Loss Type & Hinge & Vanilla & Vanilla
        \\
        Adversarial Loss Strength & 0.8 & 0.5 & 0.5
        \\
        \bottomrule
    \end{tabular}
    \label{tab:supp_implementation_details}
\end{table}

\section{More Discussions}
\subsection{Impact of the number of codebook}
We examine how the number of sub-codebooks affects reconstruction quality under a fixed total capacity. We keep the total codebook capacity (total codewords~$=4096$) and the overall latent dimension ($C=32$) constant. Four configurations are compared, ranging from a single monolithic codebook ($1\times 4096$, dim~$32$) to eight sub-codebooks ($8\times 512$, dim~$4$). The per-group dimension shrinks proportionally as the group count increases, while the sum of codewords across groups remains unchanged.

As reported in Table~\ref{tab:supp_num_code}, reconstruction quality improves consistently with more sub-codebooks, confirming that partitioning the latent space into lower-dimensional subspaces makes nearest-neighbor lookup more reliable.

\begin{table}[!htb]
    \centering
    \caption{Ablation on number of sub-codebooks.}
    \begin{tabular}{c | c c c c}
        \toprule
        Num. of Codebook / Code Dim & $1\times4096$ / 32 & $2\times2048$ / 16 & $4\times1024$ / 8 & $8\times512$ / 4
        \\
        \midrule
        rPSNR & 23.90 & 25.68 & 26.74 & 27.03
        \\
        rFID & 34.74 & 18.75 & 12.05 & 11.64
        \\
        \bottomrule
    \end{tabular}
    \label{tab:supp_num_code}
\end{table}

\subsection{Impact of different visual foundation models}
We compare several vision foundation models (VFMs) as the semantic prior extractor, including CLIP-based backbones\footnote{\url{https://github.com/mlfoundations/open_clip}}. (ConvNext, ViT-B), SigLIPv2-B, and DINOv3. As shown in Table~\ref{tab:supp_vfm_type}, stronger VFMs consistently improve perceptual quality, with ViT-B notably outperforming ConvNext under the same OpenCLIP pretraining objective. DINOv3-B delivers the best overall balance and is adopted as default, while DINOv3-L further pushes the upper bound at the cost of larger model size. These results confirm that the choice of VFM substantially impacts perceptual quality, and that the DINOv3 family provides the most robust semantic priors for our task.

\begin{table}[!htb]
    \centering
    \caption{Ablation on different Visual Foundation Models.}
    \begin{tabular}{c | c c c c}
        \toprule
        Method & PSNR & LPIPS & CLIPIQA & MANIQA
        \\
        \midrule
        ConvNext (OpenCLIP) & 25.82 & 0.2586 & 0.6321 & 0.5554
        \\
        ViT-B (OpenCLIP) & 25.91 & 0.2511	& 0.6515 & 0.5910
        \\
        SigLIPv2-B & 25.93 & 0.2473 & 0.6688 & 0.5892
        \\
        DINOv3-B & 26.16 & 0.2477 & 0.6924 & 0.6069
        \\
        DINOv3-L & 26.33 & 0.2325 & 0.7059 & 0.6351
        \\
        \bottomrule
    \end{tabular}
    \label{tab:supp_vfm_type}
\end{table}

\subsection{Limitations of HiTokSR}
\label{sec:limitations}
While HiTokSR achieves strong performance across diverse real-world benchmarks, we identify two notable limitations.
First, under severe degradation conditions such as extremely low-quality inputs with heavy degradations, the generator may occasionally produce structurally plausible but semantically inaccurate textures. This suggests that the current semantic guidance mechanism, while effective for moderate degradations, may benefit from stronger degradation-robust priors or iterative refinement in extreme cases.
Second, HiTokSR exhibits degraded performance on small text and fine-grained details, where characters may become blurred or distorted during reconstruction. This is a known challenge in generative super-resolution. We leave the investigation of these challenges to future work.

\subsection{Broader Impacts}
\label{sec:impacts}
Our method effectively improves the visual quality of real-world images, benefiting applications such as historical photo restoration and telephoto imaging. However, like other advanced generative models, it carries a potential risk of being misused to synthesize misleading high-fidelity content. We advocate for the responsible use of SR technologies and encourage the development of robust AI-generated image detection tools.

\section{More Visual Results}
\begin{figure*}[!htb]
    \centering
    \begin{subfigure}[c]{0.21\textwidth}
        \centering
        \includegraphics[width=\linewidth]{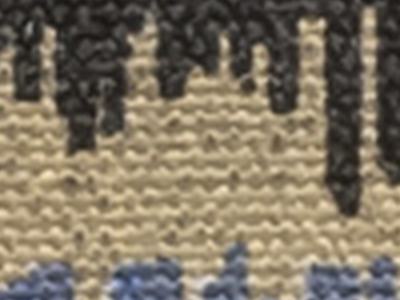}
        \subcaption*{LQ}
    \end{subfigure}
    \begin{subfigure}[c]{0.21\textwidth}
        \centering
        \includegraphics[width=\linewidth]{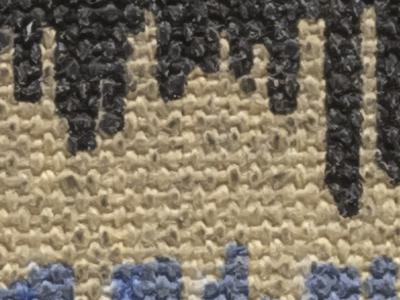}
        \subcaption*{HR}
    \end{subfigure}
    \begin{subfigure}[c]{0.21\textwidth}
        \centering
        \includegraphics[width=\linewidth]{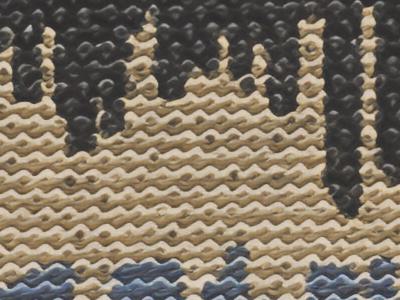}
        \subcaption*{StableSR}
    \end{subfigure}
    \begin{subfigure}[c]{0.21\textwidth}
        \centering
        \includegraphics[width=\linewidth]{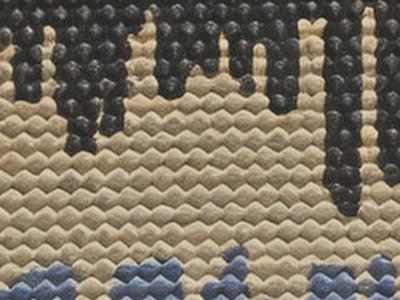}
        \subcaption*{OSEDiff}
    \end{subfigure}
    \\
    \begin{subfigure}[c]{0.21\textwidth}
        \centering
        \includegraphics[width=\linewidth]{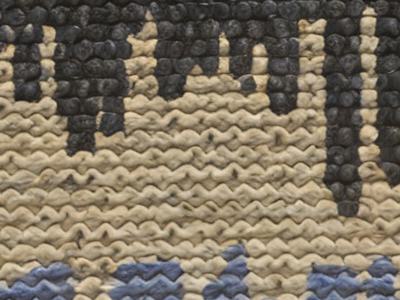}
        \subcaption*{AdcSR}
    \end{subfigure}
    \begin{subfigure}[c]{0.21\textwidth}
        \centering
        \includegraphics[width=\linewidth]{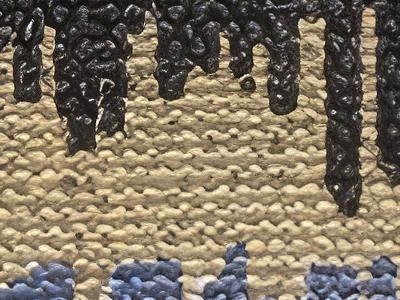}
        \subcaption*{TVQ-RAP}
    \end{subfigure}
    \begin{subfigure}[c]{0.21\textwidth}
        \centering
        \includegraphics[width=\linewidth]{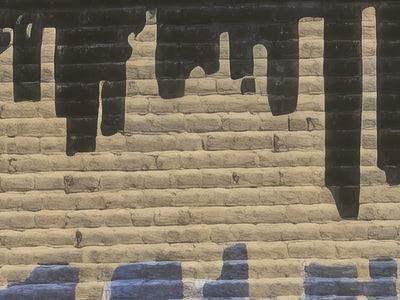}
        \subcaption*{VARSR}
    \end{subfigure}
    \begin{subfigure}[c]{0.21\textwidth}
        \centering
        \includegraphics[width=\linewidth]{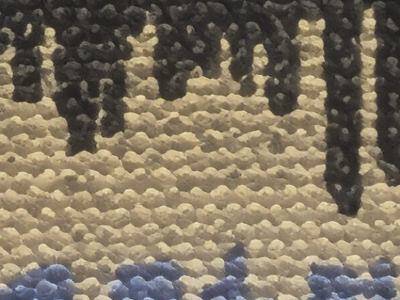}
        \subcaption*{HiTokSR}
    \end{subfigure}

    \begin{subfigure}[c]{0.21\textwidth}
        \centering
        \includegraphics[width=\linewidth]{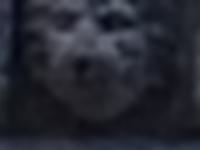}
        \subcaption*{LQ}
    \end{subfigure}
    \begin{subfigure}[c]{0.21\textwidth}
        \centering
        \includegraphics[width=\linewidth]{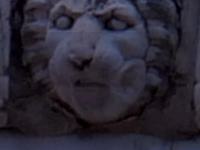}
        \subcaption*{HR}
    \end{subfigure}
    \begin{subfigure}[c]{0.21\textwidth}
        \centering
        \includegraphics[width=\linewidth]{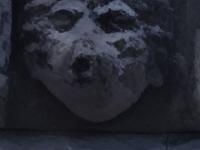}
        \subcaption*{StableSR}
    \end{subfigure}
    \begin{subfigure}[c]{0.21\textwidth}
        \centering
        \includegraphics[width=\linewidth]{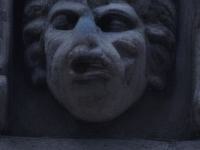}
        \subcaption*{OSEDiff}
    \end{subfigure}
    \\
    \begin{subfigure}[c]{0.21\textwidth}
        \centering
        \includegraphics[width=\linewidth]{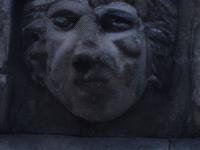}
        \subcaption*{AdcSR}
    \end{subfigure}
    \begin{subfigure}[c]{0.21\textwidth}
        \centering
        \includegraphics[width=\linewidth]{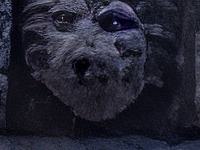}
        \subcaption*{TVQ-RAP}
    \end{subfigure}
    \begin{subfigure}[c]{0.21\textwidth}
        \centering
        \includegraphics[width=\linewidth]{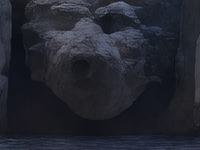}
        \subcaption*{VARSR}
    \end{subfigure}
    \begin{subfigure}[c]{0.21\textwidth}
        \centering
        \includegraphics[width=\linewidth]{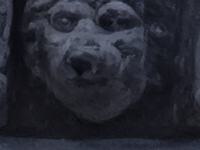}
        \subcaption*{HiTokSR}
    \end{subfigure}
    
    \begin{subfigure}[c]{0.21\textwidth}
        \centering
        \includegraphics[width=\linewidth]{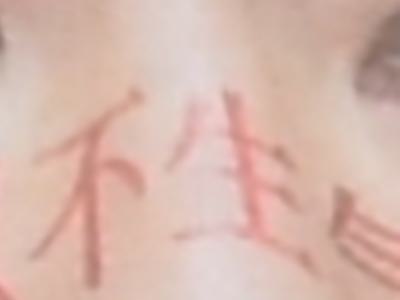}
        \subcaption*{LQ}
    \end{subfigure}
    \begin{subfigure}[c]{0.21\textwidth}
        \centering
        \includegraphics[width=\linewidth]{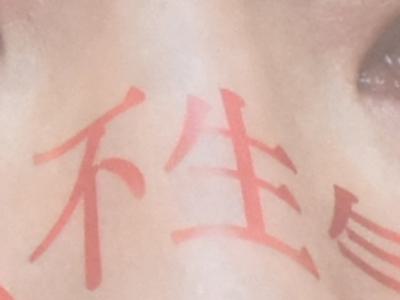}
        \subcaption*{HR}
    \end{subfigure}
    \begin{subfigure}[c]{0.21\textwidth}
        \centering
        \includegraphics[width=\linewidth]{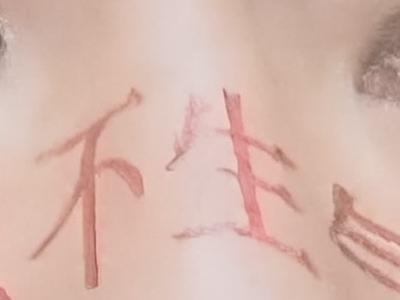}
        \subcaption*{StableSR}
    \end{subfigure}
    \begin{subfigure}[c]{0.21\textwidth}
        \centering
        \includegraphics[width=\linewidth]{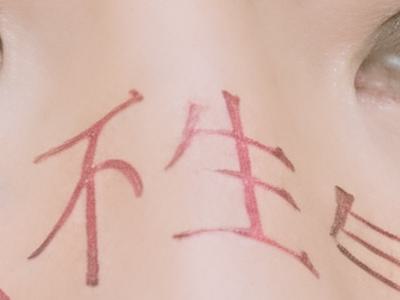}
        \subcaption*{OSEDiff}
    \end{subfigure}
    \\
    \begin{subfigure}[c]{0.21\textwidth}
        \centering
        \includegraphics[width=\linewidth]{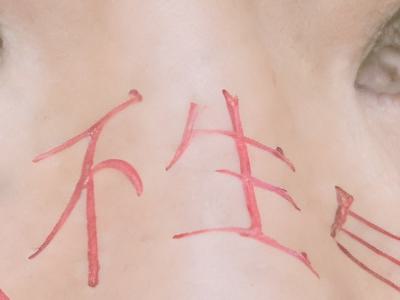}
        \subcaption*{AdcSR}
    \end{subfigure}
    \begin{subfigure}[c]{0.21\textwidth}
        \centering
        \includegraphics[width=\linewidth]{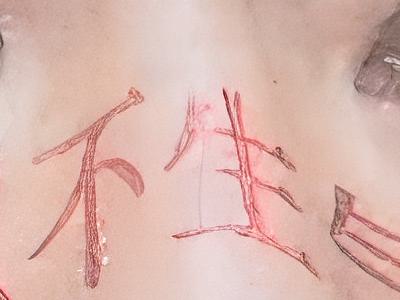}
        \subcaption*{TVQ-RAP}
    \end{subfigure}
    \begin{subfigure}[c]{0.21\textwidth}
        \centering
        \includegraphics[width=\linewidth]{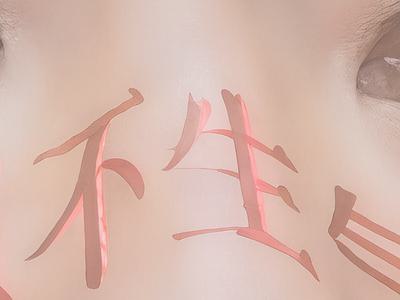}
        \subcaption*{VARSR}
    \end{subfigure}
    \begin{subfigure}[c]{0.21\textwidth}
        \centering
        \includegraphics[width=\linewidth]{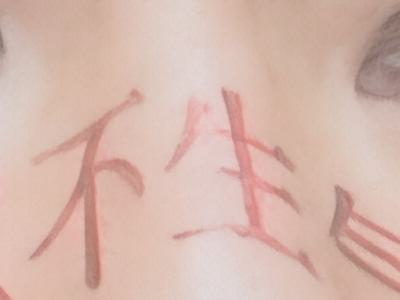}
        \subcaption*{HiTokSR}
    \end{subfigure}

    \begin{subfigure}[c]{0.21\textwidth}
        \centering
        \includegraphics[width=\linewidth]{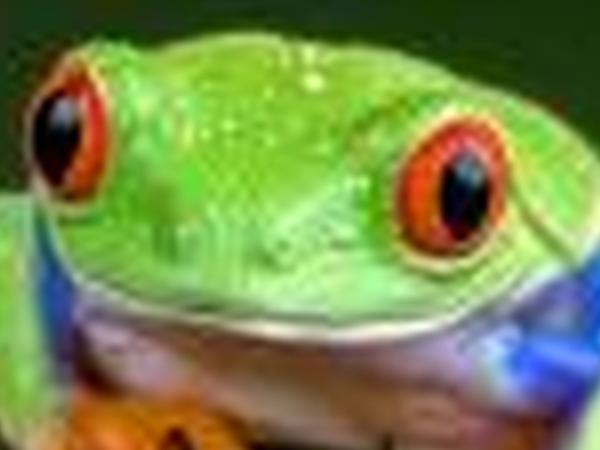}
        \subcaption*{LQ}
    \end{subfigure}
    \begin{subfigure}[c]{0.21\textwidth}
        \centering
        \includegraphics[width=\linewidth]{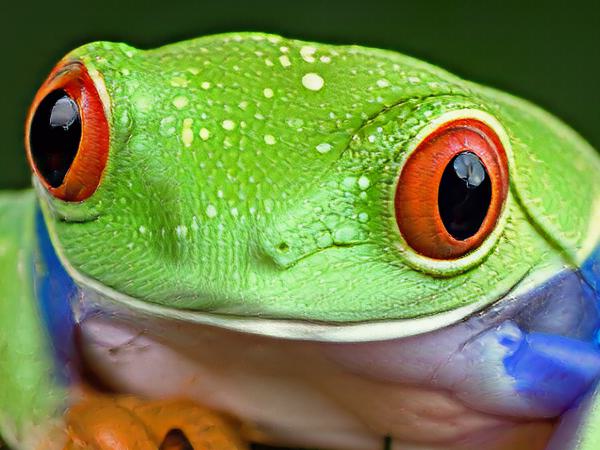}
        \subcaption*{TSD-SR}
    \end{subfigure}
    \begin{subfigure}[c]{0.21\textwidth}
        \centering
        \includegraphics[width=\linewidth]{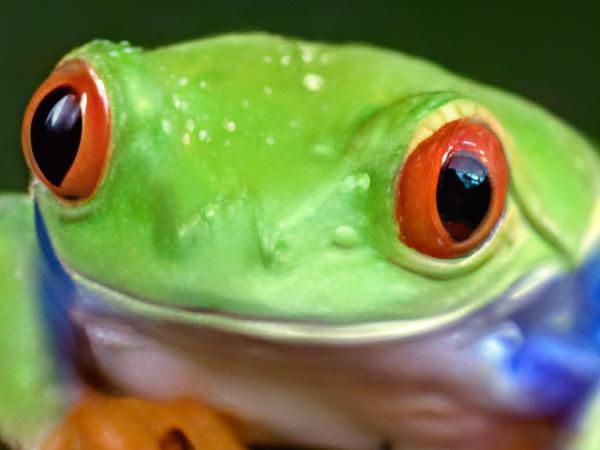}
        \subcaption*{StableSR}
    \end{subfigure}
    \begin{subfigure}[c]{0.21\textwidth}
        \centering
        \includegraphics[width=\linewidth]{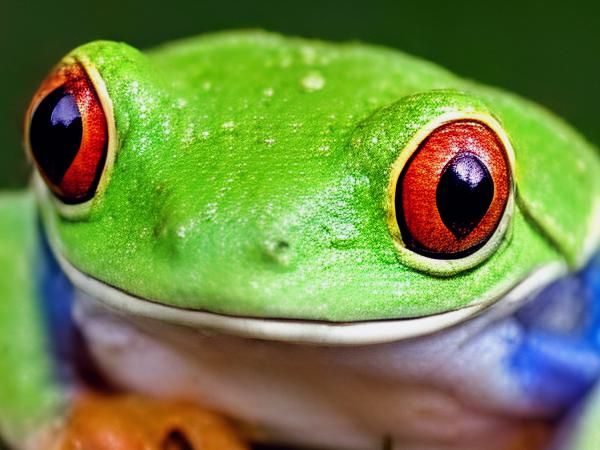}
        \subcaption*{OSEDiff}
    \end{subfigure}
    \\
    \begin{subfigure}[c]{0.21\textwidth}
        \centering
        \includegraphics[width=\linewidth]{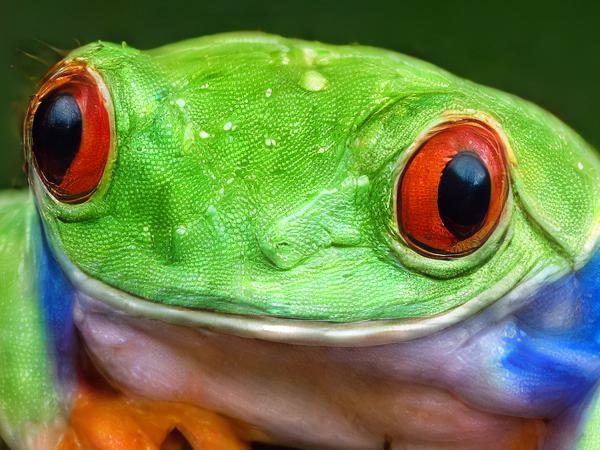}
        \subcaption*{AdcSR}
    \end{subfigure}
    \begin{subfigure}[c]{0.21\textwidth}
        \centering
        \includegraphics[width=\linewidth]{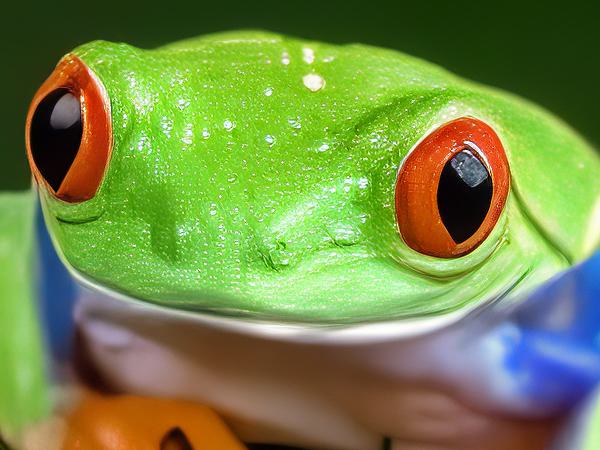}
        \subcaption*{TVQ-RAP}
    \end{subfigure}
    \begin{subfigure}[c]{0.21\textwidth}
        \centering
        \includegraphics[width=\linewidth]{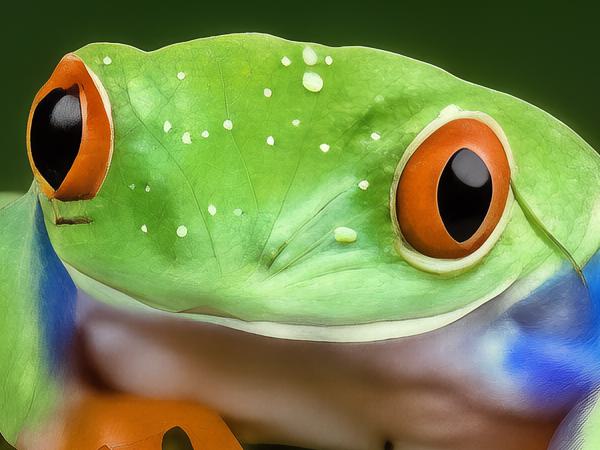}
        \subcaption*{VARSR}
    \end{subfigure}
    \begin{subfigure}[c]{0.21\textwidth}
        \centering
        \includegraphics[width=\linewidth]{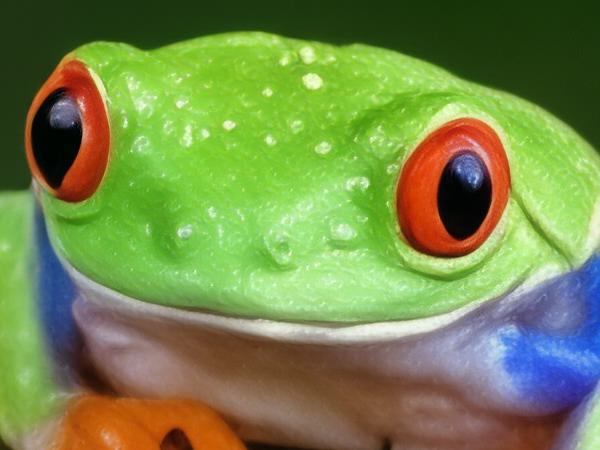}
        \subcaption*{HiTokSR}
    \end{subfigure}

    \caption{More visual comparisons with Real-SR methods on real-world benchmarks. Please zoom in for a better view.}
    \label{fig:supplementary_visual_comp_a}
\end{figure*}

\begin{figure*}[!htb]
    \centering
    \begin{subfigure}[c]{0.21\textwidth}
        \centering
        \includegraphics[width=\linewidth]{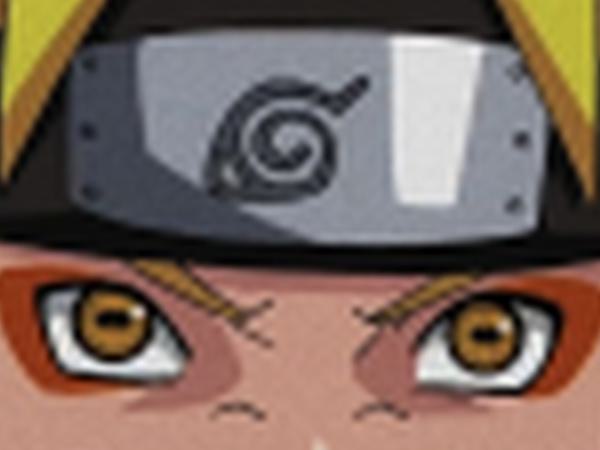}
        \subcaption*{LQ}
    \end{subfigure}
    \begin{subfigure}[c]{0.21\textwidth}
        \centering
        \includegraphics[width=\linewidth]{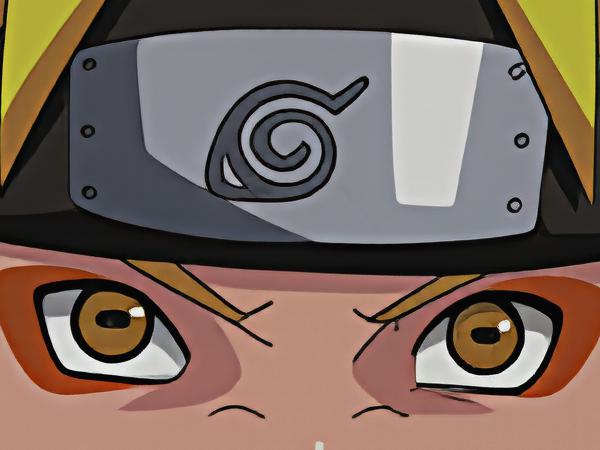}
        \subcaption*{TSD-SR}
    \end{subfigure}
    \begin{subfigure}[c]{0.21\textwidth}
        \centering
        \includegraphics[width=\linewidth]{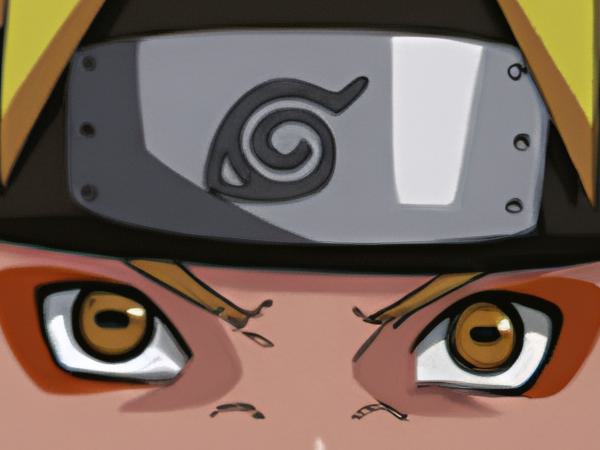}
        \subcaption*{StableSR}
    \end{subfigure}
    \begin{subfigure}[c]{0.21\textwidth}
        \centering
        \includegraphics[width=\linewidth]{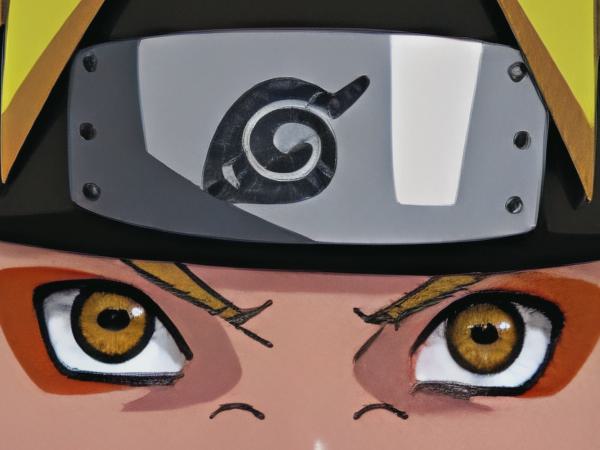}
        \subcaption*{OSEDiff}
    \end{subfigure}
    \\
    \begin{subfigure}[c]{0.21\textwidth}
        \centering
        \includegraphics[width=\linewidth]{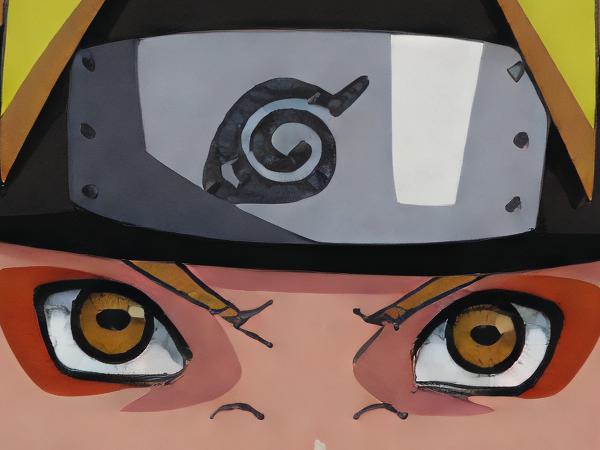}
        \subcaption*{AdcSR}
    \end{subfigure}
    \begin{subfigure}[c]{0.21\textwidth}
        \centering
        \includegraphics[width=\linewidth]{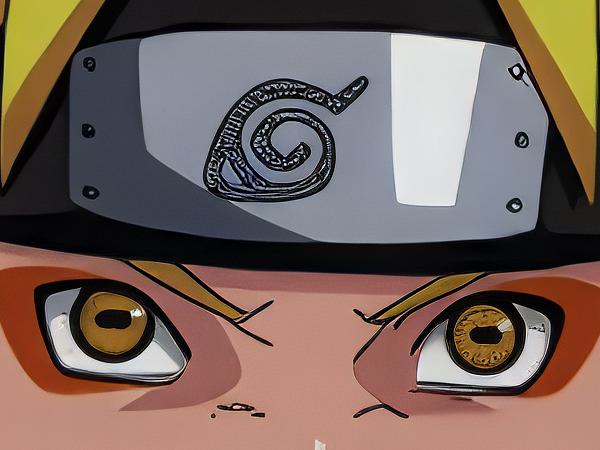}
        \subcaption*{TVQ-RAP}
    \end{subfigure}
    \begin{subfigure}[c]{0.21\textwidth}
        \centering
        \includegraphics[width=\linewidth]{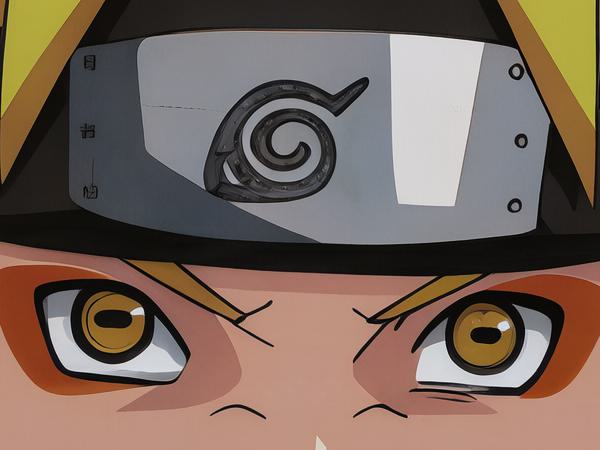}
        \subcaption*{VARSR}
    \end{subfigure}
    \begin{subfigure}[c]{0.21\textwidth}
        \centering
        \includegraphics[width=\linewidth]{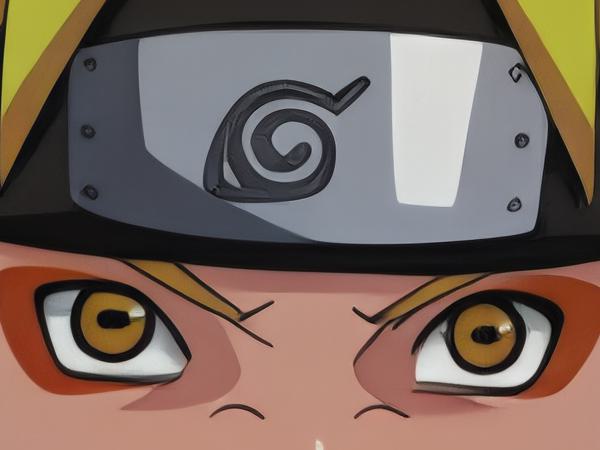}
        \subcaption*{HiTokSR}
    \end{subfigure}

    \begin{subfigure}[c]{0.21\textwidth}
        \centering
        \includegraphics[width=\linewidth]{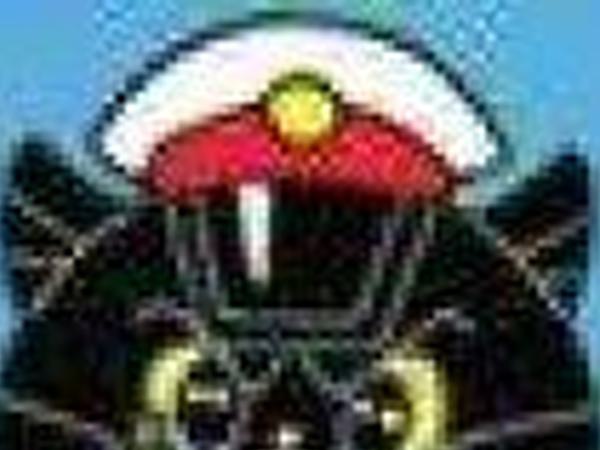}
        \subcaption*{LQ}
    \end{subfigure}
    \begin{subfigure}[c]{0.21\textwidth}
        \centering
        \includegraphics[width=\linewidth]{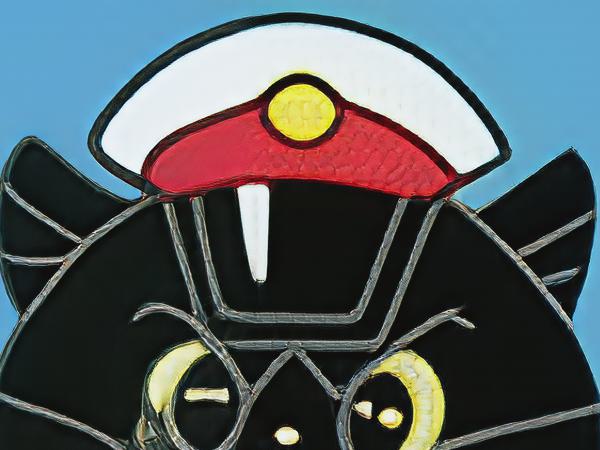}
        \subcaption*{TSD-SR}
    \end{subfigure}
    \begin{subfigure}[c]{0.21\textwidth}
        \centering
        \includegraphics[width=\linewidth]{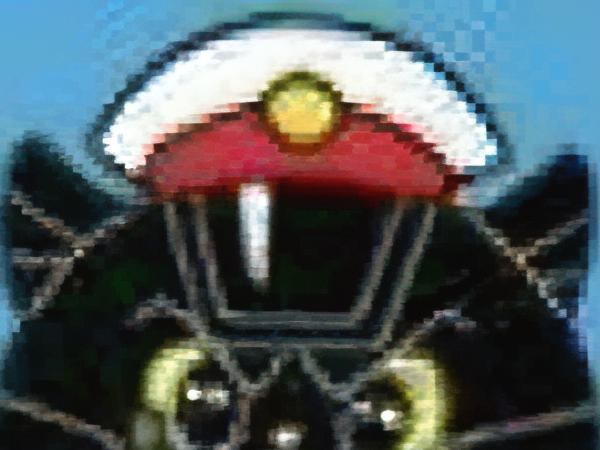}
        \subcaption*{StableSR}
    \end{subfigure}
    \begin{subfigure}[c]{0.21\textwidth}
        \centering
        \includegraphics[width=\linewidth]{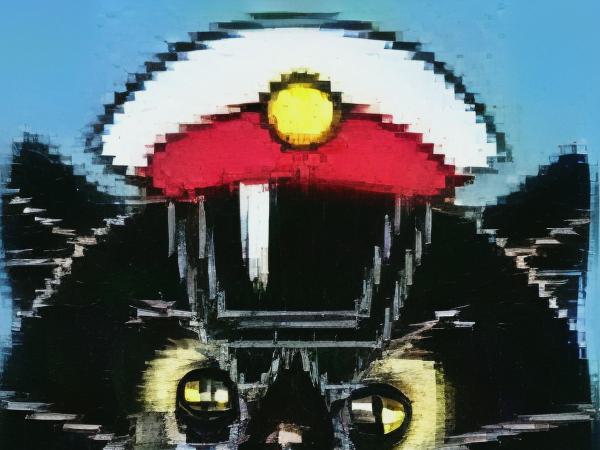}
        \subcaption*{OSEDiff}
    \end{subfigure}
    \\
    \begin{subfigure}[c]{0.21\textwidth}
        \centering
        \includegraphics[width=\linewidth]{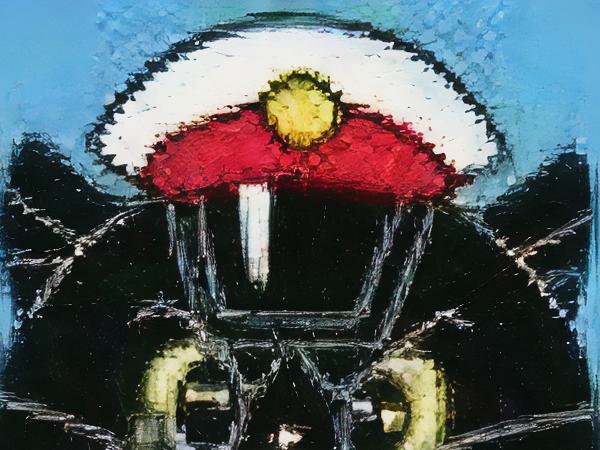}
        \subcaption*{AdcSR}
    \end{subfigure}
    \begin{subfigure}[c]{0.21\textwidth}
        \centering
        \includegraphics[width=\linewidth]{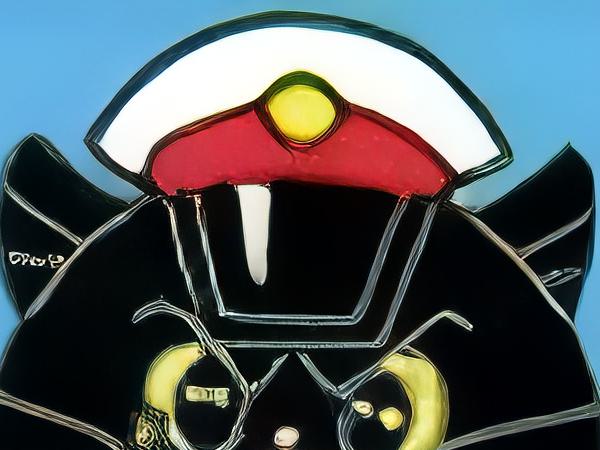}
        \subcaption*{TVQ-RAP}
    \end{subfigure}
    \begin{subfigure}[c]{0.21\textwidth}
        \centering
        \includegraphics[width=\linewidth]{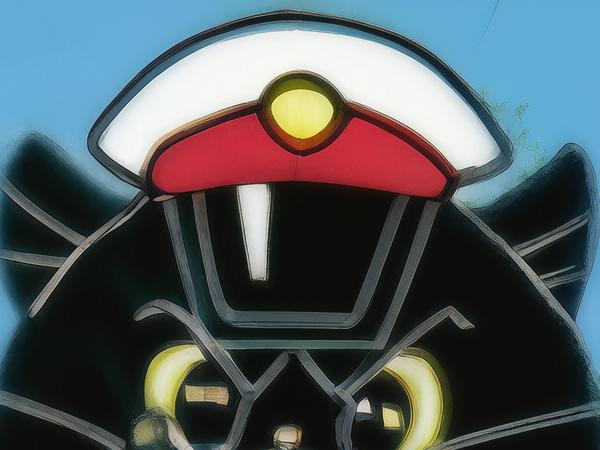}
        \subcaption*{VARSR}
    \end{subfigure}
    \begin{subfigure}[c]{0.21\textwidth}
        \centering
        \includegraphics[width=\linewidth]{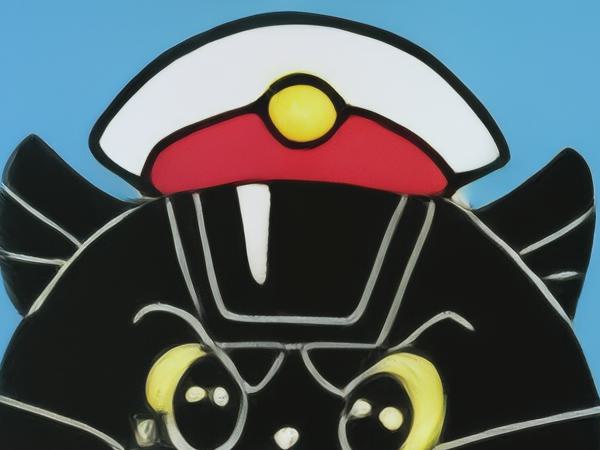}
        \subcaption*{HiTokSR}
    \end{subfigure}

    \begin{subfigure}[c]{0.21\textwidth}
        \centering
        \includegraphics[width=\linewidth]{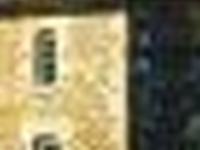}
        \subcaption*{LQ}
    \end{subfigure}
    \begin{subfigure}[c]{0.21\textwidth}
        \centering
        \includegraphics[width=\linewidth]{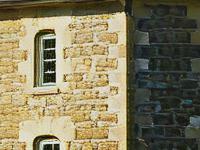}
        \subcaption*{TSD-SR}
    \end{subfigure}
    \begin{subfigure}[c]{0.21\textwidth}
        \centering
        \includegraphics[width=\linewidth]{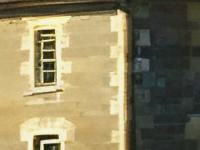}
        \subcaption*{StableSR}
    \end{subfigure}
    \begin{subfigure}[c]{0.21\textwidth}
        \centering
        \includegraphics[width=\linewidth]{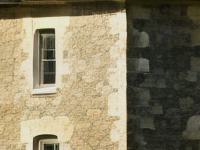}
        \subcaption*{OSEDiff}
    \end{subfigure}
    \\
    \begin{subfigure}[c]{0.21\textwidth}
        \centering
        \includegraphics[width=\linewidth]{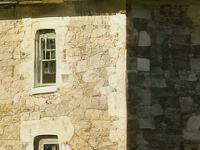}
        \subcaption*{AdcSR}
    \end{subfigure}
    \begin{subfigure}[c]{0.21\textwidth}
        \centering
        \includegraphics[width=\linewidth]{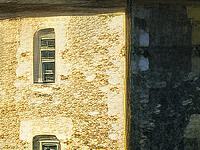}
        \subcaption*{TVQ-RAP}
    \end{subfigure}
    \begin{subfigure}[c]{0.21\textwidth}
        \centering
        \includegraphics[width=\linewidth]{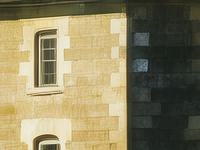}
        \subcaption*{VARSR}
    \end{subfigure}
    \begin{subfigure}[c]{0.21\textwidth}
        \centering
        \includegraphics[width=\linewidth]{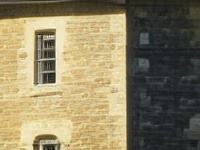}
        \subcaption*{HiTokSR}
    \end{subfigure}

    \begin{subfigure}[c]{0.21\textwidth}
        \centering
        \includegraphics[width=\linewidth]{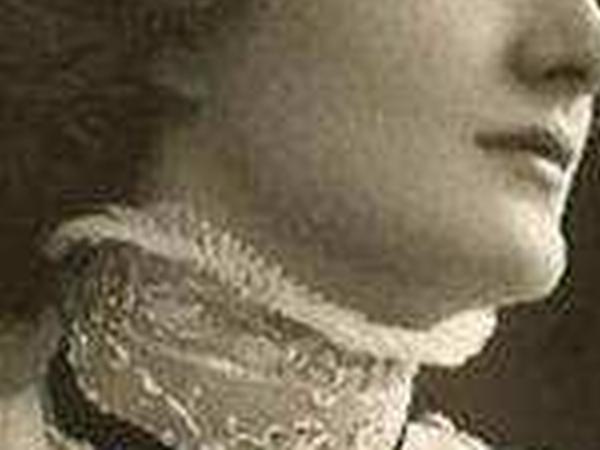}
        \subcaption*{LQ}
    \end{subfigure}
    \begin{subfigure}[c]{0.21\textwidth}
        \centering
        \includegraphics[width=\linewidth]{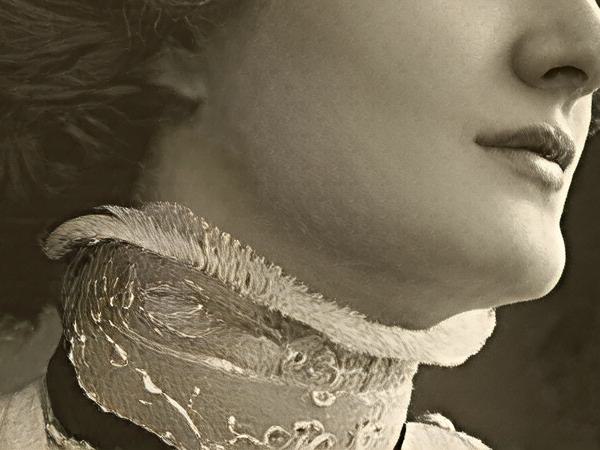}
        \subcaption*{TSD-SR}
    \end{subfigure}
    \begin{subfigure}[c]{0.21\textwidth}
        \centering
        \includegraphics[width=\linewidth]{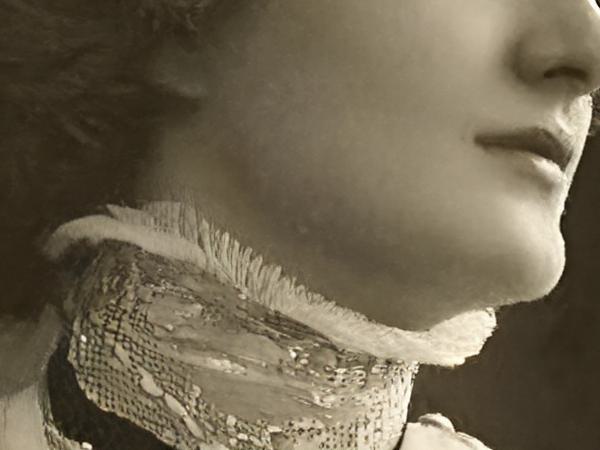}
        \subcaption*{StableSR}
    \end{subfigure}
    \begin{subfigure}[c]{0.21\textwidth}
        \centering
        \includegraphics[width=\linewidth]{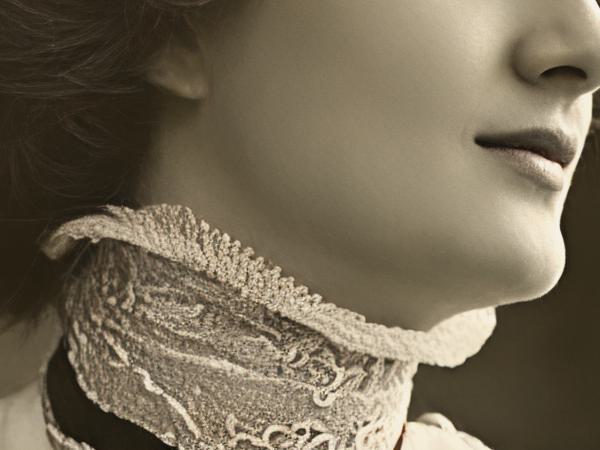}
        \subcaption*{OSEDiff}
    \end{subfigure}
    \\
    \begin{subfigure}[c]{0.21\textwidth}
        \centering
        \includegraphics[width=\linewidth]{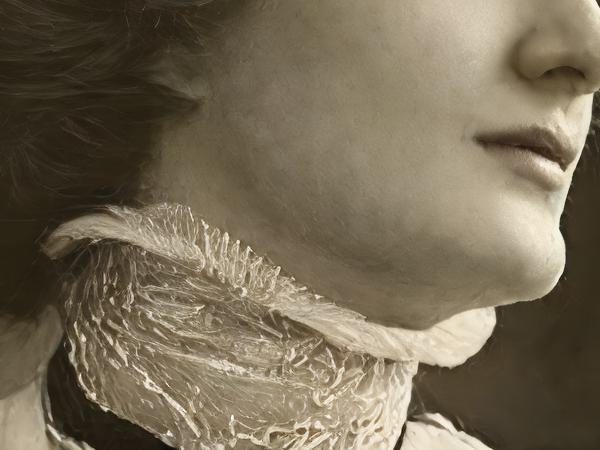}
        \subcaption*{AdcSR}
    \end{subfigure}
    \begin{subfigure}[c]{0.21\textwidth}
        \centering
        \includegraphics[width=\linewidth]{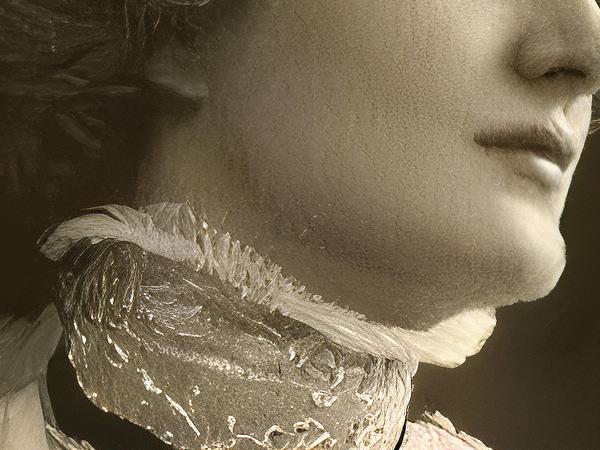}
        \subcaption*{TVQ-RAP}
    \end{subfigure}
    \begin{subfigure}[c]{0.21\textwidth}
        \centering
        \includegraphics[width=\linewidth]{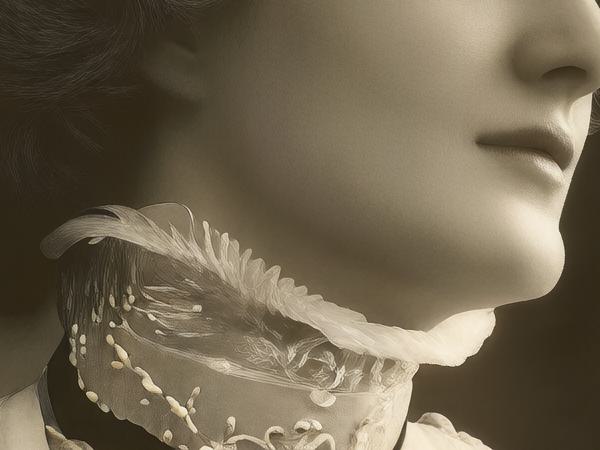}
        \subcaption*{VARSR}
    \end{subfigure}
    \begin{subfigure}[c]{0.21\textwidth}
        \centering
        \includegraphics[width=\linewidth]{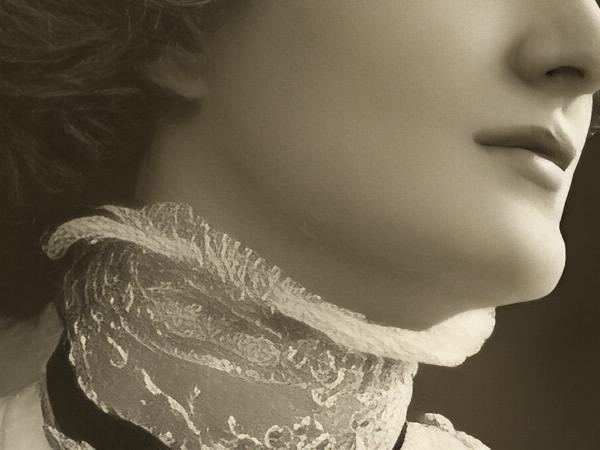}
        \subcaption*{HiTokSR}
    \end{subfigure}
    \caption{More visual comparisons with Real-SR methods on real-world benchmarks. Please zoom in for a better view.}
    \label{fig:supplementary_visual_comp_b}
\end{figure*}

We provide more visual results of HiTokSR compared with recent state-of-the-art methods in \Cref{fig:supplementary_visual_comp_a} and \Cref{fig:supplementary_visual_comp_b}.

\clearpage

\end{document}